\documentclass{article}

\usepackage{microtype}
\usepackage{graphicx}
\usepackage{subfigure}
\usepackage{booktabs} 
\usepackage[colorlinks,
            linkcolor=blue,
            anchorcolor=red,
            citecolor=blue
            ]{hyperref}  

\usepackage[preprint]{neurips_2021}

\title{On the Validity of Modeling SGD with Stochastic Differential Equations (SDEs)}

\usepackage{url}            
\usepackage{booktabs}       
\usepackage{amsfonts}       
\usepackage{nicefrac}       
\usepackage{amsmath,amsthm,amssymb}
\setcitestyle{authoryear}

\usepackage{cleveref}

\usepackage{algorithm}
\usepackage{algorithmic}
\usepackage{bbm}
\usepackage{thmtools}
\usepackage{enumitem}
\usepackage{times}
\usepackage{comment}

\usepackage{tikz-cd}
\usepackage{adjustbox}
\usepackage{thmtools} 
\usepackage{thm-restate}
\usepackage{soul}

\usepackage{tikz} 
\usetikzlibrary{shapes.misc}
\newcommand*\cancel[2][thin]{\tikz[baseline] \node [strike out,draw,anchor=text,inner sep=0pt,text=black,#1]{#2};}

\theoremstyle{plain}
\newtheorem{theorem}{Theorem}[section]
\newtheorem{lemma}[theorem]{Lemma}
\newtheorem{remark}[theorem]{Remark}
\newtheorem{corollary}[theorem]{Corollary}
\theoremstyle{definition}
\newtheorem{definition}[theorem]{Definition}

\newtheorem{example}[theorem]{Example}

\newcommand{\zhiyuan}[1]{{\color{red}{[[Zhiyuan: #1]]}}}
\newcommand{\sadhika}[1]{{\color{blue}{[[Sadhika: #1]]}}}

\newcommand{\hx}{\hat{x}} 
\newcommand{\ito}{{It\^{o}}}
\newcommand{\levy}{{L\'{e}vy}}

\usepackage{amsmath,amsfonts,bm}

\def\Secref#1{Section~\ref{#1}}

\def\eqref#1{equation~\ref{#1}}

\def\1{\bm{1}}

\makeatletter
\newcommand{\ve}{\@ifnextchar\bgroup{\velong}{{\bm{e}}}}
\newcommand{\velong}[1]{{\bm{#1}}}
\makeatother

\DeclareMathAlphabet{\mathsfit}{\encodingdefault}{\sfdefault}{m}{sl}
\SetMathAlphabet{\mathsfit}{bold}{\encodingdefault}{\sfdefault}{bx}{n}

\def\gN{{\mathcal{N}}}

\def\gS{{\mathcal{S}}}
\def\gT{{\mathcal{T}}}

\def\sN{{\mathbb{N}}}

\newcommand{\E}{\mathbb{E}}
\newcommand{\R}{\mathbb{R}}

\DeclareMathOperator{\Tr}{Tr}

\renewcommand{\R}{\mathbb{R}}
\renewcommand{\E}{\mathbb{E}}
\newcommand{\N}{\mathbb{N}}

\newcommand{\Ccal}{\mathcal{C}}

\newcommand{\Xtil}{\widetilde{X}}
\newcommand{\Dtil}{\widetilde{\Delta}}
\newcommand{\tr}{\mathrm{Tr}}

\renewcommand{\N}{\mathbb{N}}

\newcommand{\Loss}{\mathcal{L}}

\newcommand{\norm}[1]{\left\| #1 \right\|}

\newcommand{\dd}{\textup{\textrm{d}}}

\newcommand{\myparagraph}[1]{\textbf{#1}~~}

\def\vB{{\bm{B}}}

\newcommand{\bgamma}{{\bar{\gamma}}}

\def\eqref#1{(\ref{#1})}

\renewcommand{\tr}[1]{\mathrm{Tr}\left[ #1 \right]}

\newcommand{\inp}[2]{\left\langle #1,#2\right\rangle}
\usepackage{hyperref}
\usepackage{url}
\usepackage{graphicx}
\usepackage[multiple]{footmisc}

\author{%
  Zhiyuan Li\footnotemark[1]\quad Sadhika Malladi\footnotemark[1]\quad Sanjeev Arora\footnotemark[1]\textsuperscript{\ \ ~, $\dagger$}\\
  \footnotemark[1]\ \ \ Princeton University\quad \footnotemark[2]\ \ \ Institute for Advanced Study\\
  \texttt{\{zhiyuanli,smalladi,arora\}@cs.princeton.edu} 
}

\begin{document}
\maketitle
\begin{abstract}

It is generally recognized that finite learning rate (LR), in contrast to infinitesimal LR, is important for good generalization in real-life deep nets. 
Most attempted explanations propose approximating finite-LR SGD with {\ito} Stochastic Differential Equations (SDEs), but formal justification for this approximation (e.g., \citep{li2019stochastic}) only applies to SGD with tiny LR. 
Experimental verification of the approximation appears computationally infeasible. 
The current paper clarifies the picture with the following contributions: 
(a) An efficient simulation algorithm SVAG that provably converges to the conventionally used {\ito} SDE approximation. 
(b) A theoretically motivated testable necessary condition for the SDE approximation and its most famous implication, the linear scaling rule \citep{goyal2017accurate}, to hold.
(c) Experiments using this simulation to demonstrate that the previously proposed SDE approximation can meaningfully capture the training and generalization properties of common deep nets. 

\end{abstract}

\vspace{-0.4cm}
\section{Introduction}
\vspace{-0.15cm}

Training with Stochastic Gradient Gescent (SGD)~\eqref{eq:sgd_iter} and finite learning rate (LR) is largely considered essential for getting best performance out of deep nets: using infinitesimal LR (which turns the process into {\em Gradient Flow} (GF)) or finite LR with full gradients results in noticeably worse test error despite sometimes giving better training error~\citep{wu2020noisy,smith2020generalization,bjorck2018understanding}. 

Mathematical explorations of the implicit bias of finite-LR SGD toward good generalization have focused on the {\em noise} arising from gradients being estimated from small batches. 
This has motivated modeling SGD as a {\em stochastic process} and, in particular, studying Stochastic Differential Equations (SDEs) to understand the evolution of net parameters. 

Early attempts to analyze the effect of noise try to model it as as a fixed Gaussian~\citep{jastrzkebski2017three,mandt2017stochastic}.
Current approaches approximate SGD using a parameter-dependent noise distribution that match the first and second order moments of of the SGD~(\Cref{eq:motivating_sde}).
 It is important to realize that this approximation is {\em heuristic} for finite LR, meaning it is not known whether the two trajectories actually track each other closely. 
 Experimental verification seems difficult because simulating the (continuous) SDE requires full gradient/noise computation over suitably fine time intervals.   
 Recently,~\cite{li2017stochastic,li2019stochastic} provided a rigorous proof that the trajectories are arbitrarily close in a natural sense, but the proof needs the LR of SGD to be an unrealistically small (unspecified) constant so the approximation remains heuristic.

Setting aside the issue of {\em correctness} of the SDE approximation, there is no doubt it has yielded important insights of practical importance, especially the {\em linear scaling rule} (LSR; see \Cref{def:lsr}) relating batch size and optimal LR, which  allows much faster training using high parallelism~\citep{krizhevsky2014one,goyal2017accurate}. However, since the scaling rule depends upon the validity of the SDE approximation, it is not mathematically  understood when the rule fails. (Empirical investigation, with some intuition based upon analysis of simpler models, appears in~\citep{smith2020generalization,goyal2017accurate}.)

This paper casts new light on the SDE approximation via the following contributions:

\setlist{nolistsep}
\begin{enumerate}[noitemsep]
	\item A new and efficient numerical method, \emph{Stochastic Variance Amplified Gradient (SVAG)}, to test if the trajectories of SGD and its corresponding SDE are close for a given model, dataset, and hyperparameter configuration. In~\Cref{thm:sde_svag}, we prove (using ideas similar to ~\cite{li2019stochastic}) that SVAG provides an order-1 weak approximation to the corresponding SDE. (\Cref{sec:svag})
	\item \vspace*{0.2cm}Empirical testing showing that the trajectory under SVAG converges and closely follows SGD, suggesting (in combination with the previous result) that the SDE approximation can be a meaningful approach to understanding the implicit bias of SGD in deep learning.
	\item \vspace*{0.2cm}New theoretical insight into the observation in  \citep{goyal2017accurate,smith2020generalization} that linear scaling rule fails at large LR/batch sizes (\Cref{sec:failure_mode_1}). It applies to networks that use normalization layers ({\em scale-invariant} nets in \cite{arora2018theoretical}), which includes most popular architectures.  We give a necessary condition for the SDE approximation to hold: \emph{at equilibrium, the gradient norm must be smaller than its variance}. 
\end{enumerate}

\section{Preliminaries and Overview}
We use $|\cdot|$ to denote the $\ell_2$ norm of a vector and $\otimes$ to denote the tensor product. Stochastic Gradient Descent (SGD) is often used to solve optimization problems of the form
	 $\min_{x\in \R^d}  \Loss(x) := \E_\gamma \Loss_\gamma(x)$
where $\{ \Loss_\gamma : \gamma \in \Gamma \}$ is a family of functions from $\R^d$ to $\R$ and $\gamma$ is a $\Gamma$-valued variable, e.g., denoting a random batch of training data.
We consider the general case of an expectation over arbitrary index sets and distributions. 
\begin{align}
	x_{k+1} = x_{k} - \eta \nabla \Loss_{\gamma_k}(x_k), \qquad \text{\em (SGD)}
	\label{eq:sgd_iter}
\end{align}
where each $\gamma_k$ is an i.i.d. random variable with the same distribution as $\gamma$. Taking learning rate (LR) $\eta$ toward $0$ turns SGD into (deterministic) Gradient Descent (GD) with infinitesimal LR, also called {\em Gradient Flow}.  
Infinitesimal LR is more compatible with traditional calculus-based analyses, but SGD with finite LR yields the best generalization properties in practice.
Stochastic processes give a way to (heuristically) model SGD as a continuous-time evolution (i.e., stochastic differential equation or SDE) without ignoring the crucial role of noise. 
Driven by the intuition  that the benefit SGD depends primarily on the  covariance of noise in gradient estimation (and not, say, the higher moments), researchers arrived at following SDE 
for parameter vector $X_t$:  
\begin{align}
	\dd X_t = - \nabla \Loss(X_t) \dd t +  {(\eta\Sigma(X_t))}^{1/2} \dd W_t \qquad \text{\em (SDE approximation)}
	\label{eq:motivating_sde}
\end{align}
where $W_t$ is Wiener Process, and $\Sigma(X) := \E [ (\nabla \Loss_{\gamma}(X) - \nabla \Loss(X)) {(\Loss_{\gamma}(X) - \nabla \Loss(X))}^\top]$ is the covariance of the gradient noise.  
When the gradient noise is modeled by white noise as above, it is called an {\em {\ito} SDE}. Replacing $W_t$ with a more general distribution with stationary and independent increments (i.e., a \emph{{\levy} process}, described in \Cref{def:levy_process}) yields a {\em {\levy} SDE}. 

The SDE view---specifically, the belief in key role played by noise covariance---motivated the famous {\em Linear Scaling Rule}, a rule of thumb to train models with large minibatch sizes (e.g., in highly parallel architectures) by changing LR proportionately, thereby preserving the scale of the gradient noise.
\begin{definition}[Linear Scaling Rule (LSR)]~\citep{krizhevsky2014one,goyal2017accurate}\label{def:lsr}
	When multiplying the minibatch size  by $\kappa >0$, multiply the learning rate (LR) also by $\kappa$.
\end{definition}

If the SDE approximation accurately captures the SGD dynamics for a specific training setting, then LSR should work; however, LSR can work even when the SDE approximation fails.
We hope to (1) understand when and why the SDE approximation can fail and (2) provide provable and practically applicable guidance on when LSR can fail.
Experimentally verifying if the SDE approximation is valid is computationally challenging, because it requires repeatedly computing the full gradient and the noise covariance at each iteration, e.g. the \emph{Euler-Maruyama} method~\eqref{eq:noisy_gd}, which is called \emph{Noisy Gradient Descent} in the rest of the paper.
We are not aware of any empirical verification using conventional techniques, which we discuss in more detail in \Cref{subsec:app_sde_schemes}.
\Cref{sec:svag} gives a new, tractable simulation algorithm, SVAG, and presents theory and experiments suggesting it is a reasonably good approximation to both the SDE and SGD. 

\paragraph{Formalizing closeness of two stochastic processes.}
Two stochastic processes (e.g., SGD and SDE) track each other closely if they lead to similar distributions on outcomes (e.g., trained nets). Mathematics formulates closeness of distributions in terms of expectations of suitable classes of test functions\footnote{ The discriminator net in GANs is an example of test function in machine learning.}; see \Cref{subsec:svagsde}.
The test functions of greatest interest for ML are of course train and test error. These do not satisfy formal conditions such as differentiability assumed in classical theory but can be still used in experiments (see \Cref{fig:svag_accs}). 
\Cref{sec:failure_mode_1} uses test functions such as weight norm $|x_t|$, gradient norm $|\nabla \Loss(x_t)|$ and trace of noise covariance $\Tr[\Sigma(x_t)]$  and proves a sufficient condition for the failure of SDE approximation.


Mathematical analyses of  closeness of SGD and SDE will often consider the discrete process
\begin{align}
	\hx_{k+1} = \hx_{k} - \eta \nabla \Loss(\hx_k) + \eta\Sigma^{\frac{1}{2}}(\hx_k)z_k, \qquad \text{\em (Noisy Gradient Descent/NGD)} 
	\label{eq:noisy_gd}
\end{align}
where $z_k\overset{\text{i.i.d.}}{\sim} N(0,I_d)$.
A basic step in analysis will be the following  {\em Error Decomposition}:
\begin{equation}\label{eq:noise_decomposition}
\begin{split}
	 \E g(X_{\eta k}) - \E g(x_k)
	 = & \underbrace{\left( \E g(X_{\eta k})\right)-\E g(\hx_k)) }_\text{Discretization Error}+ \underbrace{\left(\E g(\hx_k) - \E g(x_k) \right)}_\text{Gap due to non-Gaussian noise} 
\end{split}
\end{equation}

\textbf{Understanding the failure caused by discretization error:} In \Cref{sec:failure_mode_1}, a testable condition of SDE approximation is derived for scale-invariant nets (i.e. nets using normalization layers). 
This condition only involves the \emph{Noise-Signal-Ratio}, but not the shape of the noise. 
We further extend this condition to LSR and develops a method predicting the largest batch size at which LSR succeeds, which only takes a single run  with small batch size. 


\subsection{Understanding the Role of Non-Gaussian Noise}
Some works have challenged the traditional assumption that SGD noise is Gaussian.
\cite{simsekli2019tailindex, nguyen2019first} suggested that SGD noise is heavy-tailed, which \cite{zhou2020theoretically} claimed causes adaptive gradient methods to generalize better than SGD. 
\cite{xie2021diffusion} argued that the experimental evidence in \citep{simsekli2019tailindex} made strong assumptions on the nature of the gradient noise, and we furthermore prove in \Cref{sec:app_heayvtailed} that their measurement method could flag Gaussian distributions as non-Gaussian.
Below, we clarify how the Gaussian noise assumption interacts with our findings.

\textbf{Non-Gaussian noise is not essential to SGD performance.}
We provide experimental evidence in \Cref{fig:ngd_sgd_accs} and \Cref{sec:app_ngd_exps} that SGD~\eqref{eq:sgd_iter} and NGD~\eqref{eq:noisy_gd} with matching covariances achieve similar test performance on CIFAR10 ( $\sim 89\%$), suggesting that even if the gradient noise in SGD is non-Gaussian, modeling it by a Gaussian estimation is sufficient to understand generalization properties.
Similar experiments were conducted in \citep{wu2020noisy} but used SGD with momentum and BatchNorm, which prevents the covariance of NGD noise from being equal to that of SGD.
These findings confirm the conclusion in \citep{cheng2020stochastic} that differences in the third-and-higher moments in SGD noise don't affect the test accuracy significantly, though differences in the second moments do.

\textbf{LSR can work when SDE approximation fails.}
 We note that \citep{smith2020generalization} derives LSR (\Cref{def:lsr}) by assuming the {\ito} SDE approximation~\eqref{eq:motivating_sde} holds, but in fact the validity of the SDE approximation is a sufficient but not necessary condition for LSR to work. 
In \Secref{sec:counterexample}, we provide a concrete example where LSR holds for all LRs and batch sizes, but the dynamics are constantly away from the {\ito} SDE limit. 
This example also illustrates that the failure of the SDE approximation can be caused solely by non-Gaussian noise, even when there is no discretization error (i.e., the loss landscape and noise distribution are parameter-independent).

\textbf{SVAG does not require Gaussian gradient noise.}
In \Cref{sec:svag}, we present an efficient algorithm SVAG to simulate the {\ito} SDE corresponding to a given training setting.
In particular, \Cref{thm:sde_svag} reveals that SVAG simultaneously causes the discretization error and the gap by non-Gaussian noise to disappear as it converges to the SDE approximation.
From \Cref{fig:svag_accs} and \Cref{subsec:app_svag_exp}, we can observe that for vision tasks, the test accuracy of deep nets trained by SGD in standard settings stays the same when interpolating towards SDE via SVAG, suggesting that neither the potentially non-Gaussian nature of SGD noise nor the discrete nature of SGD dynamics is an essential ingredient of the generalization mystery of deep learning.





\section{Related Work}
\myparagraph{Applications of the SDE approximation in deep learning.}
One component of the SDE approximation is the gradient noise distribution.
When the noise is an isotropic Gaussian distribution (i.e., $\Sigma(X_t)\equiv I$), then the equilibrium of the SDE is the Gibbs distribution.
\cite{shi2020learning} used an isotropic Gaussian noise assumption to derive a convergence rate on SGD that clarifies the role of the LR during training.
Several works have relaxed the isotropic assumption but assume the noise is constant.
\cite{mandt2017stochastic} assumed the covariance $\Sigma(X)$ is locally constant to show that SGD can be used to perform Bayesian posterior inference.
\cite{zhu2019anisotropic} argued that when constant but anisotropic SGD noise aligns with the Hessian of the loss, SGD is able to more effectively escape sharp minima.

Recently, many works have used the most common form of the SDE approximation~\eqref{eq:motivating_sde} with parameter-dependent noise covariance.
\cite{li2020reconciling} and \cite{kunin2020neural} used the symmetry of loss (scale invariance) to derive properties of dynamics (i.e., $\Sigma(X_t) X_t =0$). \cite{li2020reconciling} further used this property to explain the phenomenon of sudden rising error after LR decay in training.
\cite{smith2020generalization} used the SDE to derive the linear scaling rule (\cite{goyal2017accurate} and \Cref{def:lsr}) for infinitesimally small LR.
\cite{xie2021diffusion} constructed a SDE-motivated diffusion model to propose why SGD favors flat minima during optimization.
\cite{cheng2020stochastic} analyzed MCMC-like continuous dynamics and construct an algorithm that provably converges to this limit, although their dynamics do not model SGD.


\myparagraph{Theoretical Foundations of the SDE approximation for SGD.}
Despite the popularity of using SDEs to study SGD, theoretical justification for this approximation has generally relied upon tiny LR~\citep{li2019stochastic, hu2019diffusion}.
\cite{cheng2020stochastic} proved a strong approximation result for an SDE and MCMC-like dynamics, but not SGD.
\cite{wu2020noisy} argued that gradient descent with Gaussian noise can generalize as well as SGD, but their convergence proof also relied on an infinitesimally small LR.

\myparagraph{LR and Batch Size.}
It is well known that using large batch size or small LR will lead to worse generalization \citep{bengio2012practical, LeCun2012efficient}. 
According to \citep{keskar2016large}, generalization is harmed by the tendency for large-batch training to converge to sharp minima, but \cite{dinh2017sharp} argued that the invariance in ReLU networks can permit sharp minima to generalize well too. 
\cite{li19learningrate} argued that the LR can change the order in which patterns are learned in a non-homogeneous synthetic dataset.
Several works~\citep{hoffer2017train,smith2018a,chaudhari2018stochastic,smith2018dont} have had success using a larger LR to preserve the scale of the gradient noise and hence maintain the generalization properties of small-batch training.
The relationship between LR and generalization remains hazy, as \citep{shallue2019measuring} empirically demonstrated that the generalization error can depend on many other training hyperparameters.

\section{Stochastic Variance Amplified Gradient (SVAG)}\label{sec:svag}

Experimental verification of the SDE approximation appears computationally intractable by traditional methods.
We provide an algorithm, \emph{Stochastic Variance Amplified Gradient} (SVAG), that efficiently simulates and provably converges to the {\ito} SDE~\eqref{eq:motivating_sde} for a given training setting (\Cref{thm:sde_svag}).
Moreover, we use SVAG to experimentally verify that the SDE approximation closely tracks SGD for many common settings (\Cref{fig:svag_accs}; additional settings in \Cref{sec:app_exp}).

\subsection{The SVAG Algorithm}
For a chosen hyperparameter $l\in\N^+$, we define
\begin{align}
	x_{k+1} = x_{k} - \frac{\eta}{l} \nabla \Loss^{l}_{\bgamma_k}(x_k),
	\label{eq:svag_iter}
\end{align}
where $ \bgamma_k = (\gamma_{k,1},\gamma_{k,2})$ with $\gamma_{k,1},\gamma_{k,2}$ sampled independently and 
\[ \Loss^{l}_{\bgamma_k}(\cdot):= \frac{1+\sqrt{2l-1}}{2} \Loss_{\gamma_{k,1}}(\cdot) + \frac{1-\sqrt{2l-1}}{2}\Loss_{\gamma_{k,2}}(\cdot).\]
SVAG is equivalent to performing SGD on a new distribution of loss functions constructed from the original distribution: the new loss function is a linear combination of two independently sampled losses $\Loss_{\gamma_{k,1}}$ and $\Loss_{\gamma_{k,2}}$, usually corresponding to the losses on two independent batches.
This ensures that the expected gradient is preserved while amplifying the gradient covariance by a factor of $l$, i.e., $\sqrt{\frac{\eta}{l}}\Sigma^{l}(x) = \sqrt{\eta} \Sigma^1(x)$, where $\Sigma^{l}(x):= \E [ (\nabla \Loss^l_{\bgamma}(x) - \nabla \Loss^l(x)) {(\Loss^l_{\bgamma}(x) - \nabla \Loss^l(x))}^\top ]$. 
Therefore, the {\ito} SDE that matches the first and second order moments is always~\eqref{eq:motivating_sde}.
We note that SVAG is equivalent to SGD when $l=1$, and both the expectation and covariance of the one-step update ($x_{k+1}-x_k$) are proportional to $1/l$, meaning the direction of the update is noisier when $l$ increases.

\begin{figure}[t]
    \centering
    \vspace{-1.0cm}
    \includegraphics[width=\linewidth]{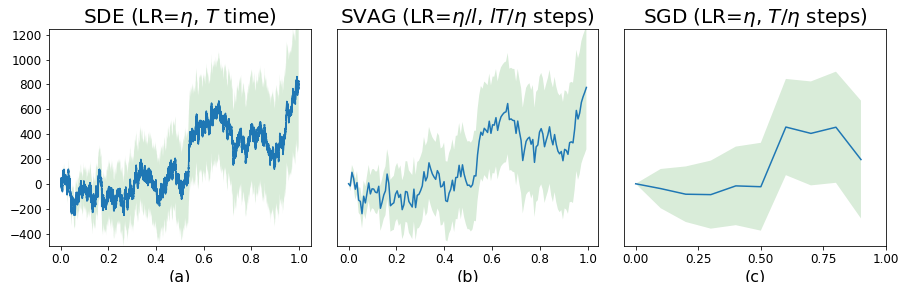}\vspace*{-0.7cm}
    \caption{\small {\ito} SDE~\eqref{eq:motivating_sde}, SVAG~\eqref{eq:svag_iter}, and SGD~\eqref{eq:sgd_iter} trajectories (blue) sampled from a distribution (green). \cite{li2019stochastic} show that $\forall T, \exists \eta$ such that SDE (a) and SGD (c) are order-1 weak approximations (\Cref{def:weak_approx}) of each other. Our result (\Cref{thm:sde_svag}) shows that $\forall T, \eta$, $\exists l$ such that SDE (a) and SVAG (b) are order-1 weak approximations of each other. In particular, \cite{li2019stochastic} requires an infinitesimal $\eta$ and our result holds for finite $\eta$.
    }\vspace{-0.3cm}
    \label{fig:proof_concept}
\end{figure}

\subsection{SVAG Approximates the SDE}
\label{subsec:svagsde}

\begin{definition}[Test Functions] Class  $G$  of continuous functions $\R^d\rightarrow\R$ has {\em polynomial growth} if $\forall g \in G$  there exist positive integers $\kappa_1,\kappa_2>0$ such that for all $x\in\R^d$,
		$| g(x) | \leq \kappa_1 (1+| x |^{2\kappa_2} ).$

	For $\alpha\in \mathbb{N}^+$, we denote by $G^\alpha$ the set of $\alpha$-times continuously differentiable functions $g$  where  all partial derivatives of form $\frac{\partial^{\overline{\alpha}} g}{\partial x^{\alpha_1}_{1}\cdots \partial x^{\alpha_d}_{d}}$ s.t. $\sum_{i=1}^d\alpha_i =\overline{\alpha}\le \alpha$, are also in $G$. 
	\label{def:test_func}
	\end{definition}
	

\begin{definition}[Order-$\alpha$ weak approximation]
Let $\{X^\eta_t:t\in[0,T]\}$ and $\{x^\eta_k\}_{k=0}^{\lfloor\frac{T}{\eta}\rfloor}$ be families of continuous and discrete stochastic processes parametrized by $\eta$. We say $\{X^\eta_t\}$ and $\{x^\eta_k\}$ are order-$\alpha$ weak approximations of each other if for every $g\in G^{2(\alpha+1)}$, there is a constant $C>0$ independent of $\eta$ such that 
\[ \max_{k=0,...,\lfloor\frac{T}{\eta}\rfloor} \left|\E g(x^\eta_k) - \E g(X^\eta_{k\eta})\right| \leq C\eta^\alpha. \]
When applicable, we drop the superscript $\eta$, and say $\{X_t\}$ and $\{x_k\}$ are {\em order-$\alpha$ (or $\alpha$ order) approximations} of each other.
\label{def:weak_approx}
\end{definition}
We now show that SVAG converges weakly to the {\ito} SDE approximation in~\eqref{eq:motivating_sde} when $l\to \infty $, i.e., $x_{lk}$ and $X_{k\eta}$ have the roughly same distribution. \Cref{fig:proof_concept} highlights the differences between our result and \citep{li2019stochastic}. \Cref{fig:svag_accs} provide verification of the below theorem, and additional settings are studied in \Cref{sec:app_exp}.
\begin{theorem}
\label{thm:sde_svag} 
	Suppose the following conditions\footnote{The $\Ccal^\infty$ smoothness assumptions can be relaxed by using the mollification technique in \cite{li2019stochastic}. } are met:
	\setlist{nolistsep}
	\begin{enumerate}[label=(\roman*), noitemsep]
		\item $\Loss \equiv \E \Loss_\gamma$ is $\Ccal^\infty$-smooth,  and $\Loss\in G^4$. 
		\item $| \nabla \Loss_\gamma (x) - \nabla \Loss_\gamma (y) |
			\leq L_\gamma| x - y |$, 
		for all $x,y \in \R^d$, where $L_\gamma>0$ is a random variable with finite moments, i.e., $\E L_\gamma^k$ is bounded for $k\in\N^+$.
		\item $\Sigma^{\frac{1}{2}}(X)$ is $\Ccal^\infty$-smooth in $X$.
	\end{enumerate}
	Let $T>0$ be a constant and $l$ be the SVAG hyperparameter~\eqref{eq:svag_iter}. Define $\{ X_t: t\in[0,T] \}$ as the stochastic process (independent of $\eta$) satisfying the {\ito} SDE~\eqref{eq:motivating_sde} and $\{x_k^{\nicefrac{\eta}{l}}: 1\le k\le \lfloor lT/\eta \rfloor\}$ as the trajectory of SVAG~\eqref{eq:svag_iter} where $x_0 = X_0$.
	Then, SVAG $\{x_k^{\nicefrac{\eta}{l}}\}$ is an order-$1$ weak approximation of the SDE $\{ X_t\}$, i.e.\,for each $g\in G^{4}$, there exists a constant $C>0$ independent of $l$ such that
	\begin{align*}
		\max_{k=0,\dots,\lfloor lT/\eta \rfloor} | \E g(x_k^{\nicefrac{\eta}{l}}) - \E g(X_{\frac{k\eta}{l}}) | \leq  Cl^{-1}.
	\end{align*}
\end{theorem}

\begin{remark}
Lipschitz conditions like (ii) are often not met by deep learning objectives. For instance using normalization schemes can make derivatives unbounded,
but if the trajectory $\{x_t\}$ stays bounded away from the origin and infinity, then (ii) holds. 
\end{remark}



\subsection{Proof Overview}
\label{sec:one_to_n_step}


 Let $\{ X^{x,s}_t : t \geq s \}$ denote the stochastic process obeying the {\ito} SDE~\eqref{eq:motivating_sde} starting from time $s$ and with the initial condition $X^{x,s}_s = x$ and  $\{ x^{x,j}_k : k \geq j \}$ denote the stochastic process (depending on $l$) satisfying SVAG~\eqref{eq:svag_iter}  with initial condition $x^{x,j}_j=x$. 
For convenience, we  define $\Xtil_{k}:=X_{\frac{k\eta}{l}}$and write $\Xtil^{x,j}_{k} := X^{x,\frac{j\eta}{l}}_{\frac{k\eta}{l}}$.  
Alternatively, we write $\Xtil_k(x,j):= \Xtil_k^{x,j}$ and $x_k(x,j):= x_k^{x,j}$.

Now for any   $1\le k \le \lfloor\frac{lT}{\eta}\rfloor$,  we interpolate between a SVAG solution $x_k$ and SDE solution $\Xtil_k$ through a series of hybrid trajectories $\Xtil_k(x_j,j)$, i.e., the weight achieved by running SVAG for the first $j$ steps  and  then SDE from time $j$ to $k$. 
The two limits of the interpolation are $\Xtil_k(x_k,k) = x_k$ (i.e., SVAG solution after $k$ steps) and $\Xtil_k(x_0,0) = \Xtil_k$ (i.e., SDE solution after $k$ time).
This yields the following error decomposition for a test function $g\in G$ (see \Cref{def:test_func}).
\begin{align*}
	 | \E g(x_k) - \E g(X_{\frac{k\eta}{l}}) |=| \E g(x_k) - \E g(\Xtil_k) |  
	 \le  \sum\nolimits_{j=0}^{k-1}\left| \E g(\Xtil_k(x_{j+1},j+1)) - \E g(\Xtil_k(x_{j},j)) \right|
\end{align*}

Note that each pair of adjacent hybrid trajectories only differ by a single step of SVAG or SDE.
We show that the one-step increments of SVAG and SDE are close in distribution along the entire trajectory by computing their moments (\Cref{lem:Dtil,lem:D}).
Then, using the Taylor expansion of $g$, we can show that the single-step approximation error from switching from SVAG to SDE is uniformly upper bounded by $O(\frac{\eta^2}{l^2})$.
Hence, the total error is $O(k\frac{\eta^2}{l^2})=O(\frac{\eta}{l})$.

\begin{restatable}[]{lemma}{lemDtil}
\label{lem:Dtil}
	Define the one-step increment of the {\ito} SDE as $\Dtil(x) = \Xtil^{x,0}_1 - x$. Then we have
	\setlist{nolistsep}
	\begin{enumerate}[label=(\roman*), noitemsep]
		\item $\E \Dtil(x)
		= -\frac{\eta}{l}{\nabla \Loss(x)} +
		\mathcal{O}(l^{-2})$,\qquad\qquad \emph{(ii)} $\E \Dtil(x)\Dtil(x)^\top
		= \frac{\eta^2}{l}  {\Sigma(x)}+ \mathcal{O}(l^{-2})$,
		\item $\E \Dtil(x)^{\otimes 3}
		= \mathcal{O}(\l^{-2})$,\qquad\qquad \qquad\qquad \quad \emph{(iv)} $\sqrt{\E | \Dtil(x)^{\otimes 4} |^2}
		= \mathcal{O}(l^{-2})$.
	\end{enumerate}
\end{restatable}

\begin{restatable}[]{lemma}{lemD}
\label{lem:D}
	Define the one-step increment of SVAG as $\Delta(x) = x^{x,0}_1 - x$. Then we have
	\setlist{nolistsep}
	\begin{enumerate}[label=(\roman*), noitemsep]
		\item $\E \Delta(x) = -\frac{\eta}{l} \nabla \Loss(x)  $,
		\item $\E \Delta(x) \Delta(x)^\top
		=  \frac{\eta^2}{l} \Sigma(x)+\frac{\eta^2}{l^2}\nabla\Loss(x) \nabla\Loss(x)^\top =\frac{\eta^2}{l}  {\Sigma(x)}+ \mathcal{O}(l^{-2})$,
		\item 
		$\E \Delta(x)^{\otimes 3}
		= \frac{\eta^3}{l^2}\frac{3-l^{-1}}{2} \Lambda(x)+\frac{\eta^3}{l^3}\left(3\overline{\nabla\Loss(x) \otimes \Sigma(x)} + \nabla\Loss(x)^{\otimes 3}\right) = \mathcal{O}(\l^{-2})$
 		\item $\sqrt{\E | \Delta(x)^{\otimes 4} |^2} = \mathcal{O}(l^{-2})$,
	\end{enumerate}
	where $\Lambda(x) :=\E {(\nabla \Loss_{\gamma_1}(x) - \nabla \Loss(x))}^{\otimes 3}$, and $\overline{\gT}$ denotes the symmetrization of tensor $\gT$, i.e., $\overline{\gT}_{ijk} = \frac{1}{6}\sum_{i',j',k'}\gT_{i'j'k'}$, where $i',j',k'$ sums over all permutation of $i,j,k$.
\end{restatable}

Though (i) and (ii) in \Cref{lem:D} hold for any discrete update with LR $=\frac{\eta}{l}$ that matches the first and second order moments of SDE~\eqref{eq:motivating_sde}, (iii) and (iv) could fail. 
For example, when decreasing LR according to LSR (\Cref{def:lsr}), even if we can use a fractional batch size and sample an infinitely divisible noise distribution, we may arrive at a different continuous limit if (iii) and (iv) are not satisfied.
(See a more detailed discussion in \Cref{subsec:app_levy_sde}) SVAG is not the unique way to ensure (iii) and (iv), and any other design (e.g. using three copies per step and with different weights) satisfying \Cref{lem:D} are also first order approximations of SDE~\eqref{eq:motivating_sde}, by the same proof.



\newcommand{\bR}{\overline{R}}
\newcommand{\bG}{\overline{G}}
\newcommand{\bN}{\overline{N}}

\newcommand{\Ri}{R_\infty}
\newcommand{\Gi}{G_\infty}
\newcommand{\Ni}{N_\infty}

\newcommand{\bRi}{\bR_\infty}
\newcommand{\bGi}{\bG_\infty}
\newcommand{\bNi}{\bN_\infty}

\section{Understanding the Failure of SDE Approximation and LSR}\label{sec:failure_mode_1}

In this section, we analyze how \emph{discretization error}, caused by large LR,  leads to the failure of the SDE approximation (\Cref{subsec:provable_sde_failure}) and LSR (\Cref{subsec:provable_lsr_failure}) for \emph{scale invariant} networks (e.g., nets equipped with BatchNorm~\citep{ioffe2015batch} and GroupNorm~\citep{wu2018group}). To get best generalization, practitioners often  add Weight Decay (WD, a.k.a $\ell_2$ regularization; see~\eqref{eq:sgd_wd}). Intriguingly, unlike the traditional setting where $\ell_2$ regularization controls the capacity of function space, for scale invariant networks, each norm ball has the same expressiveness regardless of the radius, and thus WD only regularize the model implicitly via affecting the dynamics.
 \cite{li2020reconciling} explained such phenomena by showing for training with Normalization, WD and constant LR, the parameter norm converges and  WD affects  `effective LR' by controlling the limiting value of the parameter norm.  That paper also gave experiments showing that the training loss will reach some plateau, and gave evidence of training reaching an "equilibrium" distribution that it does not get out of unless if some hyperparameter is changed.  Throughout this section we assume the existence of equilibrium for SGD and SDE.

To quantify differences in training algorithms,  we would ideally work with statistics like the train/test loss and accuracy achieved, but characterizing optimization and generalization properties of deep networks beyond the NTK regime~\citep{jacot2018ntk,allenzhu2018convergence,du2018gradient,arora2019fine,allenzhu2019learning} is in general an open problem.

Therefore, we rely on other natural test functions (\Cref{def:c_close}). 
\subsection{Failure of SDE Approximation}
\label{subsec:provable_sde_failure}
In \Cref{thm:eq_sde_sgd}, we show that the SDE-approximation of SGD is bound to fail for these scale-invariant nets  when LR gets too large. Specifically, using above-mentioned results we show that then equilibrium distributions of SGD and SDE are quite far from each other with respect to expectations of these natural test functions (\Cref{def:c_close}).

 We consider the below SDE~\eqref{eq:sde_wd} with arbitrary expected loss $\Loss(x)$ and covariance $\overline{\Sigma}(x)$, and the moment-matching SGD~\eqref{eq:sgd_wd} satisfying $ \E\Loss_\gamma(x)=\Loss(x) $ and $\overline{\Sigma}(x) = \eta\Sigma(x)$ where $ \Sigma(x)$ is the covariance of $\nabla \Loss_\gamma(x)$. In the entire \Cref{sec:failure_mode_1}, we will assume that for all $\gamma$, $\Loss_\gamma$ is \emph{scale invariant}~\citep{arora2018theoretical,Li2020An}, i.e., $\Loss_\gamma(x) = \Loss_\gamma(cx)$,  $\forall c>0$ and $x\in\R^d\setminus\{0\}$.
\begin{align}
\!\!\!\!\dd X_t\! &= \! - \nabla \big(\Loss(X_t) + \frac{\lambda}{2} |X_t|^2\big) \dd t +  {{\overline{\Sigma}}^{1/2}(X_t)} \dd W_t \label{eq:sde_wd}\\
x_{k+1} &= x_k - \eta \nabla \big(\Loss_{\gamma_k}(x_k) + \frac{\lambda}{2} |x_k|^2\big)  \label{eq:sgd_wd}
\end{align}
We will measure the closeness of two distributions by three test functions: squared weight norm $|x|^2$, squared gradient norm $|\nabla\Loss(x)|^2$, and trace of noise covariance $\Tr[\Sigma(x)]$. 
We say two equilibrium distributions are close to each other if expectations of these test functions are within a multiplicative constant.

\begin{definition}[$C$-closeness]\label{def:c_close}
Assuming the existence of the following limits, we use \\
 $R_\infty:= \lim\limits_{t\to \infty} \E|x_t|^2$, \hspace{1.40cm}$\bR_\infty:= \lim\limits_{t\to \infty} \E|X_t|^2$,\\
 $G_\infty:= \lim\limits_{t\to \infty}\E |\nabla \Loss(x_t)|^2$,\hspace{0.78cm}$\bG_\infty:= \lim\limits_{t\to \infty}\E |\nabla \Loss(X_t)|^2$,\\
 $N_\infty:= \lim\limits_{t\to \infty} \E[\Tr[\Sigma(x_t)]$, \hspace{0.46cm}$\bN_\infty:= \lim\limits_{t\to \infty} \E[\Tr[\overline{\Sigma}(X_t)]]$ \\
to denote the limiting squared norm, gradient norm and trace of covariance for SGD~\eqref{eq:sgd_wd} and   SDE~\eqref{eq:sde_wd}.  
We say the two equilibriums are \emph{$C$-close} to each other iff 
\begin{align}\label{eq:C_close}
	\frac{1}{C}\le \frac{R_\infty}{\bR_\infty},\frac{G_\infty}{\bG_\infty},\frac{ \eta N_\infty}{\bN_\infty}\le C.
\end{align}
\end{definition}
We call $\frac{\Ni}{\Gi}$ and $\frac{\bNi}{\bGi}$ the \emph{noise-to-signal ratio (NSR)}, and below we show that it plays an important role.  When the LR of SGD significantly exceeds the NSR of the corresponding SDE, the approximation fails. Of course, we lack a practical way to calculate NSR of the SDE so this result is existential rather than effective.  %
Therefore we give a  condition in terms of NSR of the SGD that {\em suffices} to imply failure of the approximation.

Experiments later in the paper show this condition is effective at showing divergence from SDE behavior.

\begin{theorem}
	\label{thm:eq_sde_sgd}
	If either (i). $\eta >  \frac{\bNi}{\bGi}(C^2-1)$ or (ii).$\frac{\Ni}{\Gi} < \frac{1}{C^2-1}$, then the equilibria of SDE~\eqref{eq:sde_wd} and SGD~\eqref{eq:sgd_wd} are not $C$-close.
\end{theorem}

\begin{remark}
Since the order-1 approximation fails for large LR, it's natural to ask if higher-order SDE approximation works. In \Cref{thm:app_1st_2nd_sde} we give a partial answer, that the same gap happens already between order-1 and order-2 SDE approximation, when $\eta\gtrsim\frac{\bNi}{\bGi}(C^2-1)$. This suggests failure of SDE approximation may be due to missing some second order term, and thus higher-order approximation in principle could avoid such failure. On the other hand, when approximation fails in such ways, e.g., increasing batch size along LSR, the performance of SGD degrades while SDE remains good. This suggests the higher-order correction term may not be very helpful for generalization. 
\end{remark}

\begin{figure*}[!t]
    \centering\vspace{-1.2cm}
    \includegraphics[width=\textwidth]{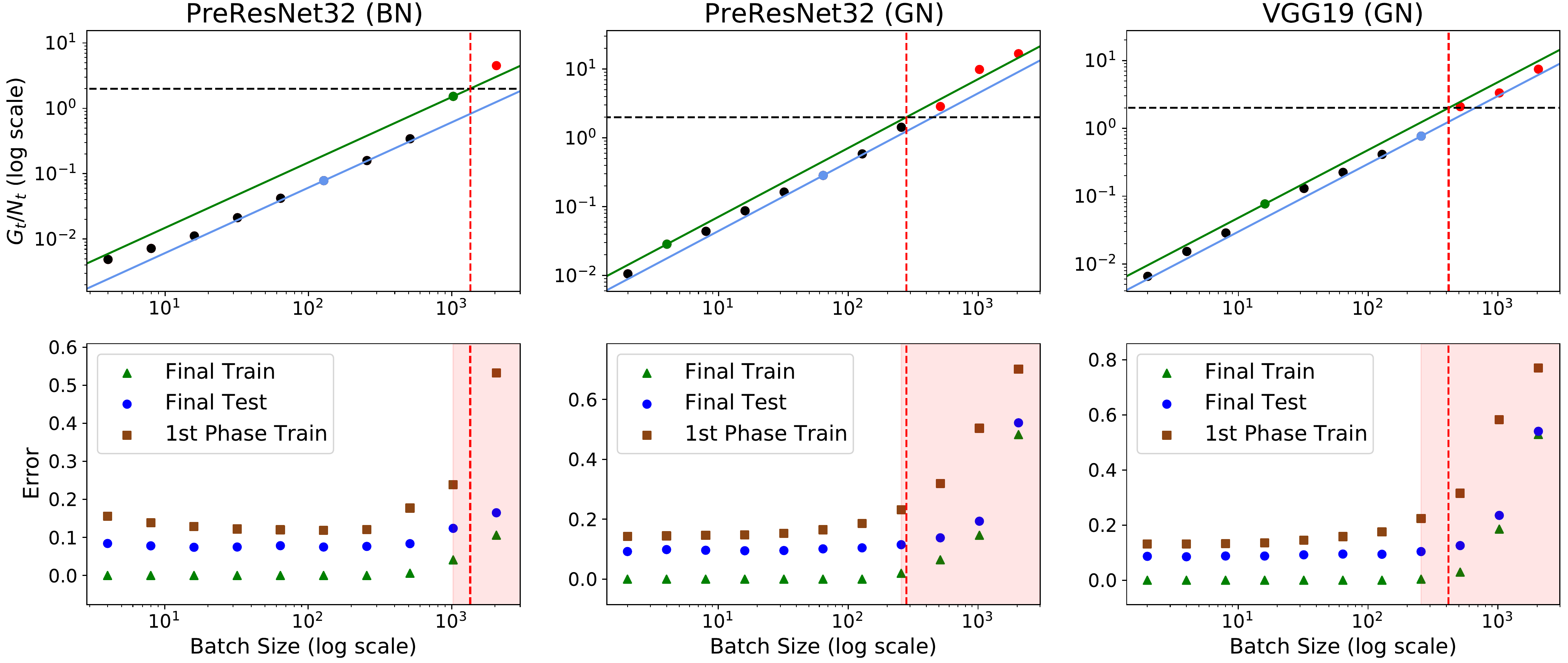}
    \vspace{-0.3cm}
    \caption{ 
\small Experimental verification for our theory on predicting the failure of Linear Scaling Rule. 
We modify PreResNet-32 and VGG-19 to be scale-invariant (according to Appendix C of \citep{li2020reconciling}).
All three settings use the same LR schedule, LR$=0.8$ initially and is decayed by $0.1$ at epoch $250$ with $300$ epochs total budget.  Here, $G_t$ and $N_t$ are the empirical estimations of $\Gi$ and $\Ni$ taken after reaching equilibrium in the first phase (before LR decay).
     Per the approximated version of \Cref{thm:lsr_kappa_bound}, i.e., $B^*=\kappa B \lesssim C^2B{\Ni^B}/{\Gi^B}$, we use baseline runs with different batch sizes $B$ to report the maximal and minimal predicted critical batch size, defined as the intersection of the threshold ($\nicefrac{G_t}{N_t}=C^2$) with the green and blue lines, respectively.
    We choose a threshold of $C^2 = 2$, and consider LSR to fail if the final test error exceeds the lowest achieved test error by more than 20\% of its value, marked by the red region on the plot. 
    Further settings and discussion are in \Cref{sec:app_exp}. 
    }\vspace{-0.cm}
    \label{fig:lsr_acc_gnr}
\end{figure*}

\subsection{Failure of Linear Scaling Rule}
\label{subsec:provable_lsr_failure}
In this section we derive a similar necessary condition for LSR to hold.

Similar to \Cref{def:c_close}, we will use $\Ri^{B,\eta},\Gi^{B,\eta},\Ni^{B,\eta}$ as test functions for equilibrium achieved by SGD~\eqref{eq:sgd_wd} when training with LR $\eta$ and mini-batches of size $B$.  
We first introduce the concept of Linear Scaling Invariance (LSI). Note here we care about the scaled ratio $\Ni^{B,\eta}/(\kappa \Ni^{\kappa B,\kappa\eta})$ because the covariance scales inversely to batch size, $\Sigma^B(x) = \kappa \Sigma^{\kappa B}(x)$.
\begin{definition}[$(C,\kappa)$-Linear Scaling Invariance]\label{def:lsi}
	We say  SGD~\eqref{eq:sgd_wd} with batch size $B$ and LR $\eta$ exhibits $(C,\kappa)$-LSI if, for a constant $C$ such that $0 < C < \sqrt{\kappa}$, 
\begin{equation}\label{eq:C_kappa_close}
    \frac{1}{C}\le \frac{\Ri^{B,\eta}}{\Ri^{\kappa B,\kappa \eta}},\frac{\Ni^{B,\eta}}{\kappa\Ni^{\kappa B,\kappa\eta}},\frac{\Gi^{B,\eta}}{\Gi^{\kappa B,\kappa \eta}}\le C.
\end{equation}
\end{definition}

We show below that $(C,\kappa)$-LSI fails if the NSR $\frac{N_\infty}{G_\infty}$ is too small, thereby giving a certificate for failure of $(C,\kappa)$-LSI even without a baseline run.

\begin{theorem}\label{thm:lsr_condition}
For any $B$, $\eta$, $C$, and $\kappa$ such that
\begin{equation}\label{eq:R_to_GN}
	\frac{\Ni^{\kappa B,\kappa\eta}}{\Gi^{\kappa B,\kappa\eta}} < (1-\frac{1}{\kappa})\frac{1}{C^2-1} -\frac{1}{\kappa},
\end{equation}
SGD with batch size $B$ and LR $\eta$ does not exhibit $(C,\kappa)$-LSI.
\end{theorem}

We now present a simple and efficient procedure to find the largest $\kappa$ for which $(C,\kappa)$-LSI will hold, providing useful guidance to make hyper-parameter tuning more efficient.
Before doing so, one must choose an appropriate value for $C$, which controls how close the test functions must be for us to consider LSR to have ``worked.''
It is an open question what value of $C$ will ensure that the two settings achieve similar test performance, but throughout our experiments across various datasets and architectures in \Cref{fig:lsr_acc_gnr} and \Cref{sec:app_exp}, we find that $C=\sqrt{2}$ works well.
One can estimate $\Gi^{B,\eta}$ and $\Ni^{B,\eta}$ from a baseline run.
Then, one can straightforwardly compute the value for the $\kappa$ threshold given in the theorem below.
We conduct this process in \Cref{fig:lsr_acc_gnr} and \Cref{sec:app_exp} to test our theory.

\begin{theorem}\label{thm:lsr_kappa_bound}
For any $B$, $\eta$, $C$, and 

\vspace{-0.5cm}

	\begin{align}\quad \kappa > C^2(1+\frac{\Ni^{B,\eta}}{\Gi^{B,\eta}}),\ \ (\approx C^2\frac{\Ni^{B,\eta}}{\Gi^{B,\eta}}\text{ when } \frac{\Ni^{B,\eta}}{\Gi^{B,\eta}}\gg 1),\end{align}
SGD with batch size $B$ and LR $\eta$ does not exhibit $(C,\kappa)$-LSI.
\end{theorem}

\section{Experiments}

\label{sec:experiments}

\begin{figure}[t]
	\vspace{-1.cm}
    \centering\vspace*{-0.2cm}
    \includegraphics[width=0.9\linewidth]{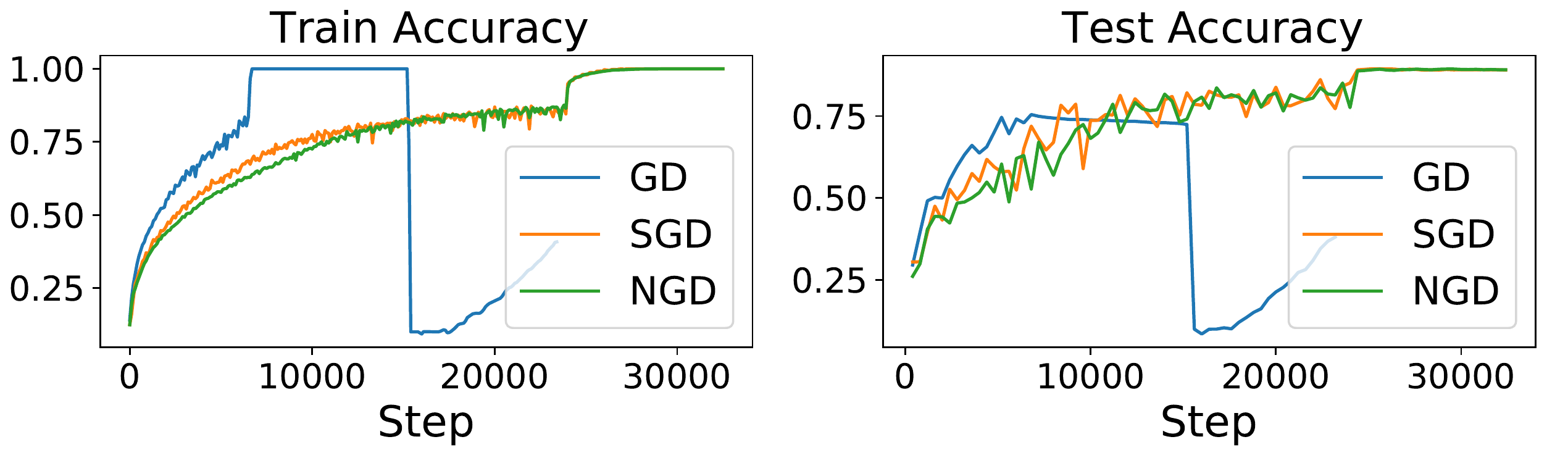}
    \vspace{-0.3cm}
    \caption{\small
    	Non-Gaussian noise is not essential to SGD performance. 
    	SGD with batch size $125$ and NGD with matching covariance have close train and test curves when training on CIFAR-10.
    	$\eta=0.8$ for all three settings and is decayed by $0.1$ at step $24000$. 
    	GD achieves 75.5\% test accuracy, and SGD and NGD achieve 89.4\% and 89.3\%, respectively. 
    	We smooth the training curve by dividing it into intervals of 100 steps and recording the average. 
    	For efficient sampling of Gaussian noise, we use GroupNorm instead of BatchNorm and turn off data augmentation. 
    	See implementation details in \Cref{sec:app_exp}.
    }
    \label{fig:ngd_sgd_accs}
\end{figure}

\Cref{fig:lsr_acc_gnr} provides experimental evidence that measurements from a single baseline run can be used to predict when LSR will break, thereby providing verification for \Cref{thm:lsr_kappa_bound}.
Surprisingly, it turns out the condition in \Cref{thm:lsr_kappa_bound} is not only sufficient but also close to necessary. 
\begin{figure}[]
    \centering
    \vspace{-0.cm}
    \includegraphics[width=0.9\linewidth]{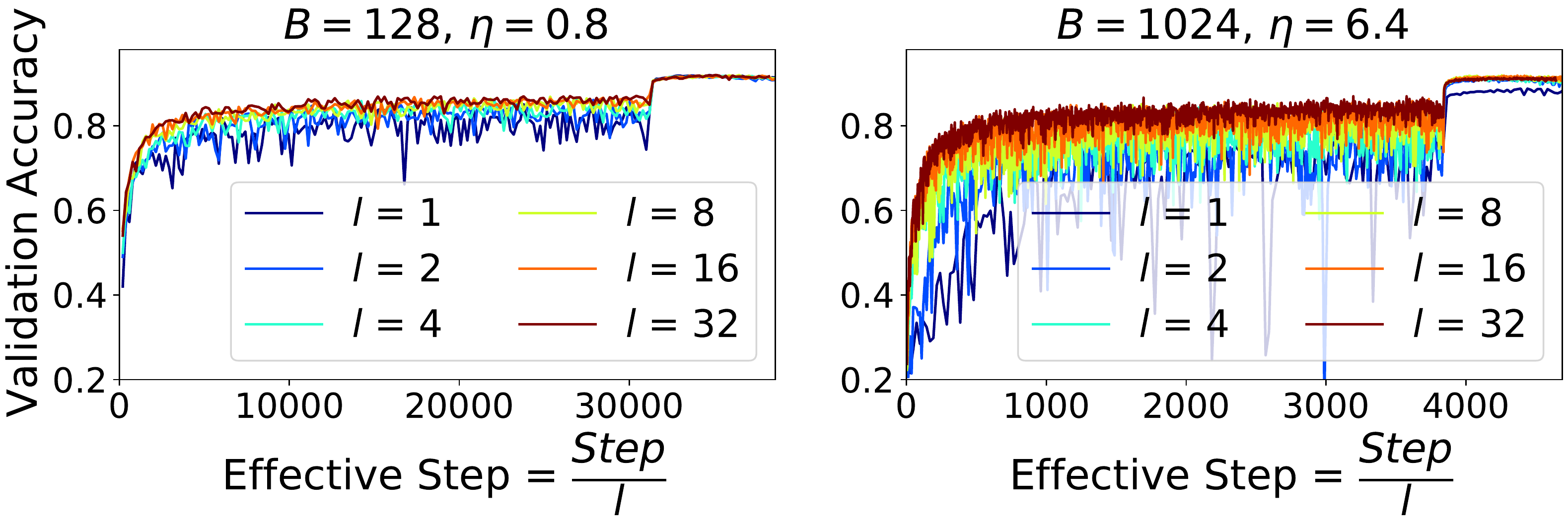}\vspace{-0.3cm}
	\caption{
		\small SVAG converges quickly and matches SGD (left) or shows the failure of the SDE approximation when LSR breaks (right).
		We train PreResNet32 with BN on CIFAR-10 for $300$ epochs, decaying $\eta$ by $0.1$ at epoch $250$.
		SVAG takes $l$ smaller steps to simulate the continuous dynamics in $\eta$ time, so we plot the accuracy against ``effective steps,'' and we note that SVAG with $l=1$ is equivalent to SGD.
		We predict in \Cref{fig:lsr_acc_gnr} that LSR (and thus, the SDE approximation) breaks at $B=1024$ for this training setting, and here we observe  SVAG converges to a limiting trajectory different from SGD, suggesting that the SDE approximation did indeed break.	}
    \label{fig:svag_accs}
\end{figure}

\Cref{fig:svag_accs} and \Cref{subsec:app_svag_exp} test SVAG on common architectures and datasets and report the results.
\Cref{thm:sde_svag} shows that SVAG converges to the SDE as $l\to\infty$, but we note that SVAG needs $l$ times as many steps as SGD to match the SDE.
Therefore, in order for SVAG to be a computationally efficient simulation of the SDE, we hope to observe convergence for small values of $l$.
This is confirmed in \Cref{fig:svag_accs} and \Cref{subsec:app_svag_exp}.
The success of SVAG in matching SGD in many cases indicates that studying the {\ito} SDE can yield insights about the behavior of SGD.
Moreover, in the case where we expect the SDE approximation to fail (e.g., when LSR fails), SVAG does indeed converge to a different limiting trajectory from the SGD trajectory.
\section{Conclusion}

We present a computationally efficient simulation SVAG (\Cref{sec:svag}) that provably converges to the canonical order-1 SDE~\eqref{eq:motivating_sde}, which we use to verify that the SDE is a meaningful approximation for SGD in common deep learning settings (\Cref{sec:experiments}).
We relate the discretization error to LSR (\Cref{def:lsr}): in \Cref{sec:failure_mode_1} we derive a testable necessary condition for the SDE approximation and LSR to hold, and in \Cref{fig:lsr_acc_gnr} we demonstrate its applicability to standard settings. 

\section*{Acknowledgement}
The authors acknowledge support from NSF, ONR, Simons Foundation, Schmidt Foundation, Mozilla
Research, Amazon Research, DARPA and SRC. ZL is also supported by Microsoft Research PhD Fellowship. 
\clearpage
\bibliographystyle{unsrt}
\bibliography{references.bib}

\appendix

\newpage
\section{Preliminaries on SDE}
\subsection{SDE Approximation Schemes}\label{subsec:app_sde_schemes}
Here, we review the common approximation schemes for SDEs and discuss why they are not efficient enough to be applied to the {\ito} SDE approximation for SGD.
We adapt the information in Chapters 13 and 14 of \cite{kloeden2011numerical}.
In general, an {\ito} SDE can be written as
\begin{equation*}
	\dd X_t = \mu(X_t, t) \dd t + \sigma(X_t,t)\dd W_t	
\end{equation*}
where $\mu$ and $\sigma$ are called the drift and diffusion coefficients respectively.
The standard {\ito} SDE~\eqref{eq:motivating_sde} used to approximate SGD sets $\mu(X_t,t) = -\nabla\Loss(X_t)$ and $\sigma(X_t,t) = (\eta\Sigma(X_t))^{1/2}$. 

Suppose we want to solve the SDE on a time interval $[0,T]$.
First, we discretize the time interval into $N$ equal steps $\tau_1,...,\tau_N$ of size $\Delta t$.
We will construct a Markov chain $Y$ that is a weak approximation in $\Delta t$ (\Cref{def:weak_approx}) to the true solution, and let $Y_0 = x_0$ where $x_0$ is the initialization for the SGD trajectory.

The \emph{Euler-Maruyama} scheme is the simplest approximation scheme, and the resulting Markov chain is an order 1 weak approximation to the true solution of the SDE. For $n\in\sN$, $0 \leq n \leq N-1$,
\begin{equation}
	Y_{n+1} = Y_n + \mu(Y_n, \tau_n)\Delta t + \sigma(Y_n, \tau_n)\Delta W_n
	\label{eq:em}
\end{equation}
where $\Delta W_n\overset{i.i.d.}{\sim}\gN(0, \Delta t)$. 
In the ML setting, computing a single step in this Markov chain requires computing the full gradient (for $\mu(Y_n,\tau_n)$) and the covariance of the gradient (for $\sigma(Y_n,\tau_n)$).
As such, modeling a single step in the recurrence requires making one pass over the entire dataset.
The error of the approximation scheme scales with $\Delta t$, so making $N$ larger (thereby requiring more recurrence steps) will improve the quality of the approximate solution.
We furthermore note that storing the gradient covariance matrix requires a large amount of memory. 
Each weight parameter in the network must be modeled by its own recurrence equation, so for modern day deep networks, this approximation seems computationally intractable.

The Euler-Maruyama scheme is considered the simplest approximation scheme for an {\ito} SDE.
A variety of other schemes, such as the Milstein and stochastic Runge-Kutta schemes, have been derived by adding a higher order corrective term, taken from the stochastic Taylor expansion, to the recurrence computation.
In particular, these schemes all still require the computation of $\mu$ and $\sigma$ at each step of the recurrence, so they remain computationally intractable for the {\ito} SDE used to approximate SGD.

\subsection{Preliminary on Stochastic Process}\label{sec:levy_sde_prelim}
\begin{definition}\label{def:levy_process}
	We call a $m$-dimensional stochastic process $X=\{X_{t}:t\geq 0\}$ a {\levy} process if it satisfies the following properties:
	\begin{itemize}
		\item $X_{0}=0$\, almost surely;
\item Independence of increments: For any $0\leq t_{1}<t_{2}<\cdots <t_{n}<\infty , X_{t_{2}}-X_{t_{1}},X_{t_{3}}-X_{t_{2}},\dots ,X_{t_{n}}-X_{t_{n-1}}$ are independent;
\item Stationary increments: For any $s<t,\ X_{t}-X_{s}$\, is equal in distribution to $ X_{t-s};$
\item Continuity in probability: For any $\varepsilon >0$ and $t\geq 0$ it holds that $\lim _{h\rightarrow 0}P(|X_{t+h}-X_{t}|>\varepsilon )=0.$
	\end{itemize}
\end{definition}

\begin{definition}\label{def:poisson_process}
	We call a counting process $\{N(t):t\geq 0\}$ a Poisson process with rate $\lambda>0$ if it satisfies the following properties:
	\begin{itemize}
		\item $N(0)=0$;
\item has independence of increments;
\item the number of events (or points) in any interval of length $t$ is a Poisson random variable with parameter (or mean) $ \lambda t$.
	\end{itemize}
\end{definition}

\section{Discussion on Non-Gaussian Noise}\label{sec:heavy_tailed}
In \Cref{sec:counterexample} we give an example where LSR holds while SDE approximation breaks.  
 In \Cref{subsec:app_levy_sde},
 we show this example to a more general setting -- infinitely divisible noise. We also explain why decreasing LR along LSR will not get a better approximation for SDE, while decreasing LR along SVAG will, since both operation preserves the same SDE approximation.  In \Cref{sec:app_heayvtailed}, we discuss the possibility where the noise is heavy-tailed and with unbounded covariance.

\subsection{LSR can hold when SDE approximation breaks}\label{sec:counterexample}
\begin{example}
Let $Z(t)$ be a $1$-dimensional Poisson process (Definition \Cref{def:poisson_process}), where $Z(t)$ follows Poisson distribution with parameter $t$. 
We assume the distribution of the gradient on single sampled data $\gamma$, $\nabla L_\gamma(x)$ is the same as $Z(1)$ for any parameter $x$. 
For a batch $\vB$ of size B (with replacement), since Poisson process has independent increments, $\nabla \Loss_\vB(x):=\frac{1}{B}\sum_{\gamma \in \vB}\nabla \Loss_\gamma(x) \overset{d}{=} \frac{Z(B)}{B}$.

Thus for any constant $T$ and initialization $x_0 = 0$, performing SGD starting from $x_0$ for $\frac{T}{B}$ steps with LR $B\eta$ and batch size $B$, the distribution of $x_{\frac{T}{B}}$ is independent of $B$, i.e., 
\begin{align*} 
x_k = x_{k-1} - B\eta \nabla\Loss_{\vB_k}(x_{k-1}) \Longrightarrow 
x_{\frac{T}{B}} \overset{d}{=}  -\eta(\underbrace{Z(B)+Z(B)+\cdots+Z(B))}_{\frac{T}{B}\text{'s }Z(B)}\overset{d}{=}-\eta Z(T). 	
\end{align*}
\end{example}

Thus LSR holds for all batch size $B$. Below we consider the corresponding NGD~\eqref{eq:noisy_gd}, $\{\hat{x}_k\}$, where

\begin{align*}
\hat{x}_k =& \hat{x}_{k-1}-B\eta \E\nabla \Loss_{\vB_k}(\hat{x}_{k-1})+ B\eta \sqrt{\E(\nabla \Loss_{\vB_k}-\E\nabla \Loss_{\vB_k})^2}z_{k-1}	\\
= & \hat{x}_{k-1} - B\eta \E\frac{Z(B)}{B} + B\eta \sqrt{\E(\frac{Z(B)}{B}-\E \frac{Z(B)}{B})^2}z_{k-1}z_{k-1}\\
=& \hat{x}_{k-1} - B\eta + \eta \sqrt{B\E(Z(1)-\E Z(1))^2}z_{k-1}\\
=& \hat{x}_{k-1} - B\eta + \eta\sqrt{B} z_{k-1},
\end{align*}
and $\{z_i\}_{i=0}^{\frac{T}{B}-1}\overset{i.i.d.}{\sim}N(0,1)$.

Thus it holds that $\hat{x}_{\frac{T}{B}} = -\eta T + \eta \sum_{k=0}^{\frac{T}{B}-1} z_{k-1} \overset{d}{=} -\eta (T + W_T)$, where $W_T$ is a Wiener process with $W_0=0$, meaning the NGD final iterate is also independent of $B$, and constant away form  the final iterate $x_{\frac{T}{B}}\overset{d}{=} -\eta Z(T)$. Indeed we can show the same result for {\ito} SDE~\eqref{eq:motivating_sde}, $\dd X_t = - \dd t + \sqrt{\eta} \dd W_t$: \[X_{\frac{T}{B}\cdot B\eta } = X_{\eta T} = \int_{t=0}^{\eta T} -\dd t+\sqrt{\eta} \dd W_t =-\eta T + \sqrt{\eta }W_{\eta T} \overset{d}{=} -\eta T + \eta W_T.\]

Thus we conclude that \textbf{LSR holds but SDE approximation fails}. Since NGD achieves the same distribution as {\ito} SDE, \textbf{the gap is solely caused by non-gaussian noise}.

However, the reader might still wonder, since batch size is always at least $1$,  there's always a lower bound for LR $\eta$ when going down along the ladder of LSR, and thus a discrete process with a finite step size of course cannot be approximated by a continuous one arbitrarily well. So isn't this example trivial? In \Cref{subsec:app_levy_sde}, we will see even if we are allowed to use fractional batch size, and thus allow $\eta\to 0$, LSR can still hold without {\ito} SDE approximation.

\subsection{Infinitely Divisible Noise and {\levy} SDE}\label{subsec:app_levy_sde}

To understand why decreasing LR along LSR will not get a better approximation for SDE, and how LSR can hold without {\ito} SDE approximation when $\eta\to 0$,  we assume the noise is infinitely divisible below for simplicity, which allows us to define SGD with fractional batch sizes and thus we can take the limit of $\eta \to 0$ along the ladder of LSR. 

That is, for the original stochastic loss $\nabla \Loss_\gamma$, for any $m\in \N^+$, there is a random loss function $\Loss^m_{\gamma'}$, such that  $\forall x\in\R^d$, the original stochastic gradient $\nabla \Loss_\gamma(x)$ is equal in distribution to the sum of $m$ i.i.d. copies of  $\nabla \Loss^m_{\gamma}(x)$:
\begin{equation}\label{eq:divisible_noise}
	\nabla\Loss_\gamma(x) \overset{d}{=} \sum_{i=1}^m\nabla\Loss^m_{\gamma'_i}(x).
\end{equation}
For SGD with batch size $B$, such a random loss function can be found when $m$ is a factor of $B$, where it suffices to define $\Loss^m$ as $m$ times the same loss with a smaller batch size $\frac{B}{m}$.\footnote{Batch loss of nets with BatchNorm is not necessarily divisble, because \eqref{eq:divisible_noise} doesn't hold, as the individual loss depends on the entire batch of data with the presence of BN. 
Still, it holds for ghost BatchNorm~\citep{hoffer2017train} with $B$ equal to the number of mini-ghost batches.
}
In other words, we can phrase LSR in a more general form, which only involves the distribution of the noise, but not the generating process of the noise (e.g. noise from sampling a batch with replacement). 

\begin{definition}[generalized Linear Scaling Rule (gLSR)]\label{def:glsr}
	Keep LR the same. Replace $\nabla \Loss_\gamma$ by $\nabla \Loss_\gamma^m$ and multiply the total number of steps by $m$.
\end{definition}

It's well known that every infinitely divisible distribution corresponds to a $d$-dimensional {\levy} process (\Cref{def:levy_process}) $Z_x(t)\in \R^d$, in the sense that $\nabla \Loss_\gamma(x) \overset{d}{=} Z_x(1)$ \cite{ken1999levy}. 
If we further assume there is a $m$-dimensional {\levy} process $Z'(t)$ and a function $\sigma(x):\R^{d}\to\R^{d\times m}$ such that for every $x$, $\nabla \Loss_\gamma(x)-\nabla \Loss(x)  \overset{d}{=}  \sigma(x)Z(1)$, then by Theorem 2.2 in \citep{protter1997euler}, SGD~\eqref{eq:levy_sde_sgd} will converge to a limiting continuous dynamic, which we denote as the \emph{{\levy} SDE}, as the LR decreases to $0$ along LSR. 
\begin{align}\label{eq:levy_sde_sgd}
	x_k = x_{k-1} -  \eta \nabla\Loss^m_{\gamma_{k-1}}(x_{k-1}), \quad  \textrm{$\forall k=1,\ldots, \lfloor \frac{Tm}{\eta}\rfloor$ }
\end{align}
Formally, $X_t$ is the solution of the following SDE driven by a {\levy} process.
	\begin{equation}\label{eq:levy_sde}
		\dd X_t  = -  \nabla \Loss(X_t)\dd t+  \eta\sigma(X_t)\dd Z_{ t/\eta}, \quad \textrm{ $\forall t\in [0,T]$}
	\end{equation}
In the special case where $Z_t$ is the $d$-dimensional Brownian motion (note $\{\eta Z_{t/\eta}\}_{t\ge 0}$ and $\{\sqrt{\eta} Z_t\}_{t\ge 0}$ have the same distributions for the sample paths), $\sigma(x)\in\mathbb{R}^{d\times d}$ will be the square root of the noise covariance, $\Sigma^{\frac{1}{2}}(x)$.

\begin{figure}[t]
\adjustbox{scale=1.3,center}{
\begin{tikzcd}[row sep=huge]
\text{NGD} \arrow[rr, shift left, color = red,"\text{SVAG}"] \arrow[rr, shift right, color = blue,"\text{LSR}"']                                                      
   &  & \text{{\ito} SDE} \arrow[d, "\text{\Large\ ?}", no head, equals, description]  \\
\text{SGD} \arrow[u, "\substack{\text{Gaussian}\\ \text{Noise}}"'] \arrow[rr, color = blue,"\text{LSR}"'] \arrow[rru, color = red, "\text{SVAG}"'] &  & \text{{\levy} SDE}  
\end{tikzcd}	
}
\caption{Taking $\eta\to 0$ and keeping the first two moments are not enough to converge to {\ito} SDE limit, e.g. decreasing LR along LSR can converge to another limit, {\levy} SDE. Red and blue arrows means taking limit of the dynamics when LR $\eta\to 0$ along the SVAG and LSR respectively. Here we assume the noise in SGD is infinitely divisible such that the LR can go to $0$ along LSR. For NGD, i.e., SGD with Gaussian noise, both SVAG and LSR (Linear Scaling Rule) approaches the same continuous limit. This does not hold for SGD with non-Gaussian noise.}	\label{fig:diagram}\vspace{-0.5cm}
\end{figure}
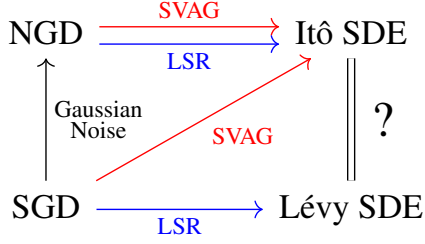

\textbf{Why decreasing LR along LSR will not get a better approximation for SDE:} The {\levy} SDE is equal to the {\ito} SDE only when the noise is strictly Gaussian.
Thus the gap induced by non-Gaussian noise will not vanish even if both SGD and NGD decrease the LR along LSR, as it will converge to the gap between {\ito} SDE and {\levy} SDE. 
See~\Cref{fig:diagram} for a summary of the relationships among SGD, NGD, {\ito} SDE, and {\levy} SDE.

Since decreasing LR along LSR converges to a different limit than SVAG does, it's natural to ask which part of the approximation in \Cref{lem:D} fails for the former.
By scrutinizing the proof of \Cref{lem:D}, we can see (i) and (ii) still hold for any stochastic discrete process with LR $\frac{\eta}{l}$ and matching first and second order moments,  while the term $\frac{\eta^3(3-l^{-1})}{2l^2}\Lambda(x)$ now becomes $\frac{\eta^3}{l}\Lambda(x)$ for SGD along LSR, which is larger by an order of $l$.
Therefore, the single-step approximation error becomes $O(l^{-1})$ and the total error after $\lfloor\nicefrac{Tl}{\eta}\rfloor$ steps remains constant.\footnote{Such error does not only occur in the third order moment. It also appears in the higher moments. Therefore simply assuming the noise distribution is symmetric (thus $\Lambda=0$) won't fix this gap.}

\textbf{SDE approximation is not necessary for LSR, even for LR $\eta\to 0$:} We also note that though \cite{smith2020generalization} derives LSR by assuming the {\ito} SDE approximation holds, this is only a sufficient but not necessary condition for LSR. 
In \Secref{sec:counterexample}, we provide a concrete example where LSR holds for all LRs and batch sizes, but the dynamics are constantly away from {\ito} SDE limit. The loss landscape and noise distribution are constant, i.e., parameter-independent. This is also an example where the gap between SGD and {\ito} SDE is solely caused by non-gaussian noise, but not the discretization error.

\subsection{Heavy-tailed Noise and Unbounded Covariance}
\label{sec:app_heayvtailed}
\cite{simsekli2019tailindex} experimentally found that the distribution of the SGD noise appears to be heavy-tailed and proposed to model it with an $\alpha$-stable process. In detail, in Figure 1 of \citep{simsekli2019tailindex}, they show that the histogram of the gradient noise computed with AlexNet on CIFAR-10 is more close to that of $\alpha$-stable random variables, instead of that of Gaussian random variables. However, a more recent paper~\citep{xie2021diffusion} pointed out a fundamental limitation of methodology in \citep{simsekli2019tailindex}: \citep{simsekli2019tailindex} made a hidden but very restrictive assumption that the noise of each parameter in the model is distributed identically. Moreover, their test (\Cref{thm:test}) of the tail-index $\alpha$ works only under this assumption. Thus the empirical measurement in \citep{simsekli2019tailindex} ($\widehat{\alpha}<2$) doesn't exclude the possibility that that stochastic gradient noise follows a joint multivariate Gaussian. 

\newcommand{\sas}{\gS\alpha\gS}
\newcommand{\one}{\bm{1}}
\begin{theorem}\label{thm:test}
 \citep{mohammadi2015estimating}
 	Let $\{X_i\}_{i=1}^K$ be a collection of i.i.d. random variables with $X_1\sim \sas(\sigma)$ and $K=K_1\times K_2$. Define $Y_i:=\sum_{j=1}^{K_1}X_{j+(i-1)K_1}$ for $i\in \{1,\ldots, K_2\}$. Then the estimator
 	\begin{align}
 		\widehat{\ \frac{1}{\alpha} \ }:=\frac{1}{\log K_1}\left( 
 		\frac{1}{K_2}\sum_{i=1}^{K_2}\log |Y_i|-\frac{1}{K}\sum_{i=1}^K\log |X_i|
 		\right).
 	\end{align}
 converges to $\frac{1}{\alpha}$ almost surely, as $K_2\to \infty$. Here $\sas(\sigma)$ is the $\alpha$-stable distribution defined by $X\sim \sas(\sigma)\Longleftrightarrow \E[\exp(i w X)] = \exp(-|\sigma w|^\alpha)$.
\end{theorem}

We provide the following theoretical and experimental evidence on vision tasks to support the argument in \cite{xie2021diffusion} that it is reasonable to model the stochastic gradient noise by joint Gaussian random variables instead of $\alpha$-stable random variables even \emph{for finite learning rate}. (Note SVAG (e.g., \Cref{fig:svag_accs}) only shows that when LR becomes infinitesimally small, replacing the noise by Gaussian noise gets similar performance.)
\begin{enumerate}
\item In \Cref{fig:lsr_acc_gnr}, we find that the trace of covariance of noise is bounded and the empirical average doesn't grow with the number of samples/batches (this is not plotted in the current paper). However, an $\alpha$-stable random variable has unbounded variance for $\alpha<2$.
\item In Figures \ref{fig:ngd_sgd_accs}, \ref{fig:ngd_sgd_b125}, \ref{fig:ngd_sgd_b500}, and \ref{fig:ngd_sgd_vgg}, we show directly that replacing the stochastic gradient noise by Gaussian noise with the same covariance gets almost the train/test curve and the final performance.
\item Applying the test in \Cref{thm:test} on joint multivariate Gaussian random variables can yield an estimate ranged from $1$ to $2$ for the tail-index $\alpha$, but for Gaussian variables, $\alpha=2$. (\Cref{thm:test_on_gaussian})
\end{enumerate}

Another recent work~\cite{zhang2020why} also confirmed that the noise in stochastic gradient in ResNet50 on vision tasks is finite. However, they also found the noise for BERT on Wikipedia+Books dataset could be heavy-tailed: the empirical variance is not converging even with $10^7$ samples. We left it as a future work to investigate how does SDE approximate SGD on those tasks or models with heavy-tailed noise.

\begin{theorem}\label{thm:test_on_gaussian}
	Let $K=K_1\times K_2 = d\times m\times K_2$, where $K_1,K_2,d,m\in\N^+$. Let $\{X_i\}_{i=1}^K$ be a collection of random variables where $X_{(j-1)d:jd} \overset{i.i.d.}{\sim}  N(0,\Sigma)$, for each $j\in\{1,\ldots, mK_2\}$. Then we have 
			\[
			\E\left[\frac{1}{\log K_1}\left( 
		\frac{1}{K_2}\sum_{i=1}^{K_2}\log |Y_i|-\frac{1}{K}\sum_{i=1}^K\log |X_i|
		\right)	 \right] = \frac{1}{2}\frac{\log m+ \log \one^\top\Sigma\one - \frac{1}{d} \sum_{i=1}^d \log \Sigma_{ii}}{\log m + \log d}.
			\]

Specifically, when $d=K_1$ and $m=1$, taking $\Sigma = \beta\one\one^\top + (1-\beta)I$, we have 
\[\E\left[\frac{1}{\log K_1}\left( 
		\frac{1}{K_2}\sum_{i=1}^{K_2}\log |Y_i|-\frac{1}{K}\sum_{i=1}^K\log |X_i|
		\right)	 \right] = \frac{1}{2}\frac{\log (\beta d^2 + (1-\beta)d)}{\log d},\]	
and 
\[\left\{ \frac{1}{2}\frac{\log (\beta d^2 + (1-\beta)d)}{\log d}\mid \beta\in[0,1]\right\} = [\frac{1}{2},1].\]

\end{theorem}

\begin{proof}
	\begin{align}
		&\E\left[\frac{1}{\log K_1}\left( 
		\frac{1}{K_2}\sum_{i=1}^{K_2}\log |Y_i|-\frac{1}{K}\sum_{i=1}^K\log |X_i|
		\right)	 \right]
	= \E\left[\frac{1}{\log K_1}\left( 
		\log |Y_1|-\frac{1}{d}\sum_{i=1}^d\log |X_i|
		\right)	 \right]. 
	\end{align}
	
Note $Y_1$ is gaussian with standard deviation $\sqrt{\one^\top\Sigma\one\cdot m}$ and $X_i$ is gaussian with standard deviation $\sqrt{\Sigma_{ii}}$. Thus $\E[\log|Y_1| -\log|X_i|] = \log \sqrt{\one^\top\Sigma\one\cdot m} - \log \sqrt{\Sigma_{ii}}$. Thus we have 
	\begin{align}
		&\E\left[\frac{1}{\log K_1}\left( 
		\frac{1}{K_2}\sum_{i=1}^{K_2}\log |Y_i|-\frac{1}{K}\sum_{i=1}^K\log |X_i|
		\right)	 \right]\\
	= &\E\left[\frac{1}{\log K_1}\left( 
		\log |Y_1|-\frac{1}{d}\sum_{i=1}^d\log |X_i|
		\right)	 \right]\\
	= &\frac{1}{2}\frac{\log m+ \log \one^\top\Sigma\one - \frac{1}{d} \sum_{i=1}^d \log \Sigma_{ii}}{\log m + \log d}
	\end{align}
\end{proof}

\section{Omitted Derivation in \Cref{sec:svag}}\label{appsec:svag_details}
We prove \Cref{thm:sde_svag} in this section.
The derivation is based on the following two-step process, following the agenda of \citep{li2019stochastic}:
\setlist{nolistsep}
\begin{enumerate}[noitemsep]
	\item Showing that the approximation error on a finite interval ($N = \lfloor \frac{Tl}{\eta}\rfloor$ steps) can be upper bounded by the sum of expected one-step errors.
(\Cref{thm:one_step_to_multi_steps}, which is Theorem 3 in \citep{li2019stochastic})
	\item \vspace{0.2cm}Showing the one-step approximation error of SVAG is of order $2$, and so the approximation on a finite interval is of order $1$. (\Cref{lem:Dtil,lem:D})
\end{enumerate}

\subsection{Relating one-step to $N$-step approximations}
\label{sec:one_to_n_step}

Let us consider generally the question of the relationship between one-step approximations and approximations on a finite interval. Let $T>0$  and $N=\lfloor lT/\eta \rfloor$.
Let us also denote for convenience $\Xtil_{k}:=X_{\frac{k\eta}{l}}$. Further, let $\{ X^{x,s}_t : t \geq s \}$ denote the stochastic process obeying the same \Cref{eq:motivating_sde}, but with the initial condition $X^{x,s}_s = x$. We similarly write $\Xtil^{x,j}_{k} := X^{x,\frac{j\eta}{l}}_{\frac{k\eta}{l}}$ and denote by $\{ x^{x,j}_k : k \geq j \}$ the stochastic process (depending on $l$) satisfying \Cref{eq:svag_iter} but with $x_j=x$.

Now, let us denote the one-step changes
\begin{align}
	\textrm{SVAG:}\quad \Delta(x) := x^{x,0}_1 - x ,
	\qquad\qquad
	\textrm{SDE:} \quad \Dtil(x) := \Xtil^{x,0}_1 - x.
\end{align}

\begin{figure}[htbp]
\centering\includegraphics[width=0.6\linewidth]{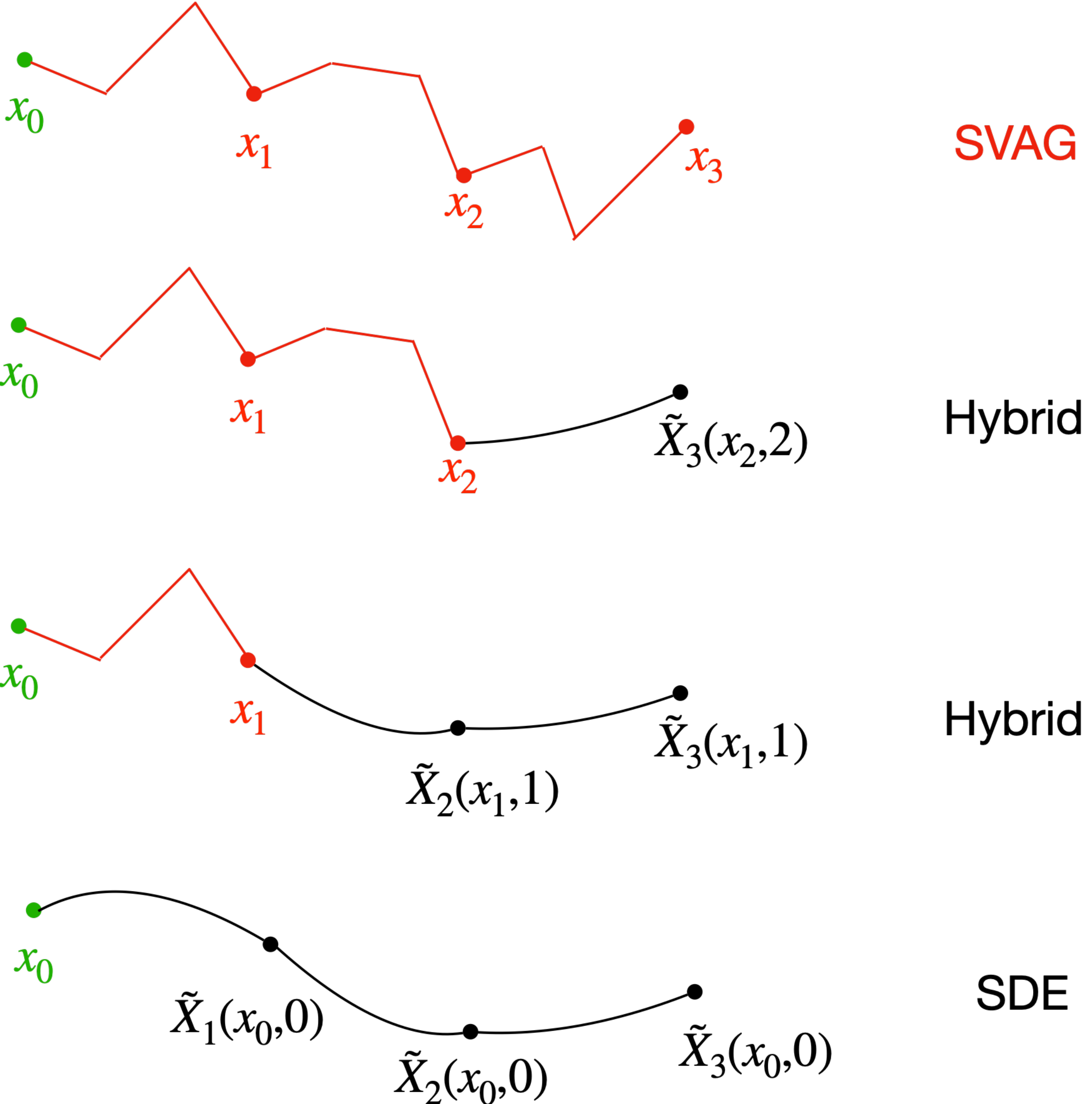}
\caption{\small	
	To relate the one-step error to the error over a finite interval, we construct interpolating hybrid trajectories between SVAG and SDE as shown in the figure. 
	Each hybrid trajectory is built by using the second to last SVAG point in the previous trajectory as the initial condition and then running the SDE for the remainder of the time interval.
}
\label{fig:app_proof_overview}
\end{figure}

The following result is adapted from \citep{li2019stochastic} to our setting, which relates one-step approximations with approximations on a finite time interval.
To prove it, we will construct hybrid trajectories interpolating between SVAG~\eqref{eq:svag_iter} and the SDE~\eqref{eq:motivating_sde}, as shown in \Cref{fig:app_proof_overview}.

\begin{restatable}[Adaption of Theorem 3 in \citep{li2019stochastic}]{theorem}{onesteptomultistep}
\label{thm:one_step_to_multi_steps}
	Suppose the following conditions hold:
	\setlist{nolistsep}
	\begin{enumerate}[label=(\roman*), noitemsep]
		\item There is a function $K_1\in G$ independent of $l$ such that
$\left|
				\E \Delta(x)^{\otimes s} - \E \Dtil(x)^{\otimes s}
			\right|
			\leq K_1(x)l^{-2}$
		for $s=1,2,3$ and $
			\sqrt{\E 
			\left|
			\Delta(x)^{\otimes 4}
			\right|^2}
			\leq K_1(x)l^{-2}.$
		\item \vspace{0.2cm} For all $m\geq 1$, the $2m$-moment of $x^{x,0}_k$ is uniformly bounded w.r.t. $k$ and $l$, i.e.\,there exists
		a $K_2 \in G$, independent of $l,k$, such that $\E | x^{x,0}_k |^{2m} \leq K_2(x)$, 
		for all $k=0,\dots, \lfloor lT/\eta \rfloor$.
	\end{enumerate}
	Then, for each $g\in G^{4}$, there exists a constant $C>0$, independent of $l$, such that\vspace{-0.7cm}
	
	\begin{align*}
	\max_{k=0,\dots,\lfloor lT/\eta \rfloor}
	\left|
	\vspace{-1cm}	\E g(x_k) - \E g(X_{\frac{k\eta}{l}})
	\right| \leq C l^{-2}
\end{align*}
\end{restatable}

\begin{proof}[Proof of \Cref{thm:one_step_to_multi_steps}]
	Let $T,l>0$,  $N=\lfloor lT/\eta \rfloor$ and for convenience we also define $\Xtil_{k}:=X_{\frac{k\eta}{l}}$. Further, let $\{ X^{x,s}_t : t \geq s \}$ denote the stochastic process obeying the same \Cref{eq:motivating_sde}, but with the initial condition $X^{x,s}_s = x$.  We similarly write $\Xtil^{x,j}_{k} := X^{x,\frac{j\eta}{l}}_{\frac{k\eta}{l}}$ and denote by $\{ x^{x,j}_k : k \geq j \}$ the stochastic process (depending on $l$) satisfying \Cref{eq:svag_iter} but with $x_j=x$. Alternatively, we write $\Xtil_k(x,j):= \Xtil_k^{x,j}$ and $x_k(x,j):= x_k^{x,j}$.  By definition, $\Xtil_k(x_k,k) = x_k$ and $\Xtil_k(x_0,0) = \Xtil_k$.

Thus we have for any $1\le k \le \lfloor\frac{lT}{\eta}\rfloor$, we can decompose the error as illustrated in \Cref{fig:app_proof_overview},
\begin{align*}
	 &| \E g(x_k) - \E g(X_{\frac{k\eta}{l}}) |=| \E g(x_k) - \E g(\Xtil_k) |  \\
	 \le & \sum_{j=0}^{k-1}\left| \E g(\Xtil_k(x_{j+1},j+1)) - \E g(\Xtil_k(x_{j},j)) \right| \\
	 \le & \sum_{j=0}^{k-1}\left| \E u^{k,j+1}(\Xtil_{j+1}(x_{j},j)) - \E u^{k,j+1}(x_{j+1}(x_{j},j)) \right| \\
	 \le & \sum_{j=0}^{k-1}\left| \E u^{k,j+1}(\Xtil_{1}(x_{j},0)) - \E u^{k,j+1}(x_{1}(x_{j},0)) \right| \\
	 \le & \sum_{j=0}^{k-1} \E\left[ \left| \E u^{k,j+1}(\Xtil_{1}(x_{j},0)) - \E u^{k,j+1}(x_{1}(x_{j},0)) \right| \big\vert x_j\right], 
\end{align*}
where $u^{k,j+1}(x)$ is defined as $\E g(X_k(x,j+1))$  and the second to the last step is because of  SDE~\eqref{eq:motivating_sde} is time-homogeneous. By Proposition~25 in \citep{li2019stochastic}, $u^{k,j+1}\in G^4$ uniformly, thus by \Cref{lem:u_eta_estimate}, we know there exists $K(x)=\kappa_1(1+|x|^{2\kappa_2})\in G$ such that 

\begin{align*}
	 | \E g(x_k) - \E g(X_{\frac{k\eta}{l}}) |
	\le  \sum_{j=0}^{k-1}\E\left[ K(x_j)l^{-2}\right]
	\le  \sum_{j=0}^{k-1}\E\left[ \kappa_1(1+|x_j|^{2\kappa_2})l^{-2}\right]
\end{align*}

By assumption (ii), we know the there is some $K' \in G$,
\begin{align*}
	 | \E g(x_k) - \E g(X_{\frac{k\eta}{l}}) |
	 \le  \sum_{j=0}^{k-1}\E\left[ \kappa_1(1+|x_j|^{2\kappa_2})l^{-2}\right]
	\le  \sum_{j=0}^{\lfloor\frac{lT}{\eta}\rfloor-1}\E\left[ \kappa_1(1+|x_j|^{2\kappa_2})l^{-2}\right]
	\le K'(x) l^{-1},
\end{align*}
which completes the proof.
\end{proof}
Recall that 
\begin{align}
	\textrm{SVAG:}\quad \Delta(x) := x^{x,0}_1 - x ,
	\qquad\qquad
	\textrm{SDE:} \quad \Dtil(x) := \Xtil^{x,0}_1 - x.
\end{align}

\begin{lemma}
\label{lem:u_eta_estimate}
	Suppose $u^1,\ldots,u^k \in G^{4}$ uniformly, that is, $u^1,\ldots,u^k \in G$ and there's a single $K_0\in G$ such that $\left|\tfrac{\partial^s u}{\partial x_{(i_1)},\dots x_{(i_j)}}(x)\right| \le K_0(x)$, for $s=1,2,3,4$ and $i_j\in \{1,2,\ldots,d\}, j\in \{1,\ldots,s\}$. Let assumption (i),(ii) in Thm.~\ref{thm:one_step_to_multi_steps} hold and $K_1(x),K_2(x)$ be such functions. Then, there exists some $K\in G$, independent of $l,r$, such that
	\begin{align*}
		\left|
			\E u^r(x^{x,0}_1) - \E u^r(\Xtil^{x,0}_1)
		\right| \leq K(x) l^{-2}
	\end{align*}
\end{lemma}
\begin{proof}
W.L.O.G, we can assume $K_0(x) = \kappa_{0,1}(1+|x|^{2\kappa_{0,2}}) \le K_0(x)^2$, for $\kappa_{0,1}>0, \kappa_{0,2}\in\N^+$, thus for $\alpha\in[0,1]$ and $x,y\in\R^d$, we have $K_0((1-\alpha)x+ \alpha y) \le \max(K_0(x),K_0(y))\le K_0(x)+K_0(y)$. We also assume $\E K_0(x_1^{x,0})^2\le K^2_2(x)$.

	Using Taylor's theorem with the Lagrange form of the remainder, we have for any $j\in\{1,\ldots,k\}$, 
	\begin{align*}
		u^r(x^{x,0}_1) - u^r(\Xtil^{x,0}_1)
		=& \sum_{s=1}^{3} \tfrac{1}{s!}
		\sum_{i_1,\dots,i_j=1}^{d}
		\prod_{j=1}^{s} [\Delta_{(i_j)}(x) - \Dtil_{(i_j)}(x)]
		\tfrac{\partial^s u^r}{\partial x_{(i_1)},\dots x_{(i_j)}}(x) \\
		+& \tfrac{1}{4!}
		\sum_{i_1,\dots,i_4=1}^{d}
		\left[ \tfrac{\partial^4 u^r}{\partial x_{(i_1)},\dots x_{(i_4)}}
			(x + a \Delta (x)) \prod_{j=1}^{4} \Delta_{(i_4)}(x)\right] \\
		-& \tfrac{1}{4!}
		\sum_{i_1,\dots,i_4=1}^{d}
		\left[ \tfrac{\partial^4 u^r}{\partial x_{(i_1)},\dots x_{(i_4)}}
			(x + a \Delta (x)) \prod_{j=1}^{4} \Delta_{(i_4)}(x)\right] 
	\end{align*}
	where $a ,\widetilde{ a }\in[0,1]$. 
	
	Taking expectations over the first term,  using assumption (i) of Thm.~\ref{sec:one_to_n_step}, we get
	\begin{align*}
	\left| \E\left[ \sum_{s=1}^{3} \tfrac{1}{s!}
		\sum_{i_1,\dots,i_j=1}^{d}
		\prod_{j=1}^{s} [\Delta_{(i_j)}(x) - \Dtil_{(i_j)}(x)]
		\tfrac{\partial^s u^r}{\partial x_{(i_1)},\dots x_{(i_j)}}(x) \right] \right|
	\le  l^{-2}(\frac{d}{1!} + \frac{d^2}{2!}+ \frac{d^3}{3!})K_1(x)K_0(x)
	\end{align*}

	Taking expectations over the second term,  using assumption (i) of \Cref{sec:one_to_n_step} and \Cref{app:lem:delta_prod_estimate}, we get
	\begin{align*}
	&\left| \E\left[ \tfrac{1}{4!}
		\sum_{i_1,\dots,i_j=1}^{d}
		\left[ \tfrac{\partial^4 u^r}{\partial x_{(i_1)},\dots x_{(i_4)}}
			(x + a \Delta (x)) \prod_{j=1}^{4} \Delta_{(i_4)}(x)\right]  \right] \right|\\
	\le & 	\tfrac{1}{4!}  
		\sum_{i_1,\dots,i_4=1}^{d}
		\E\left| \tfrac{\partial^4 u^r}{\partial x_{(i_1)},\dots x_{(i_4)}}
			(x + a \Delta (x)) \right| \left|\prod_{j=1}^{4} \Delta_{(i_4)}(x)\right| \\
	\le & 	\tfrac{1}{4!}  
		\sum_{i_1,\dots,i_4=1}^{d}
		\sqrt{\E\left| \tfrac{\partial^{(\alpha+1)} u^r}{\partial x_{(i_1)},\dots x_{(i_4)}}
			(x + a \Delta (x)) \right|^2\E \left|\prod_{j=1}^{4} \Delta_{(i_4)}(x)\right|^2 }\\
	\le & 	\tfrac{1}{4!l^2}  
		\sum_{i_1,\dots,i_4=1}^{d}
		\sqrt{\E\left|K_0(x)+K_0(x_1^{x,0}) \right|^2K_1(x)^2}\\
	\end{align*}
	
	Note that by assumption (ii) of \Cref{sec:one_to_n_step}, we have 
	\begin{align*}
	\E\left|K_0(x)+K_0(x_1^{x,0}) \right|^2	\le 2K_0(x)^2+ 2 \E K_0(x_1^{x,0})^2 \le 2K_0(x)^2+2K_2(x)^2 \le (2K_0(x)+2K_2(x))^2.
	\end{align*}

Thus, 
	\begin{align*}
	&\left| \E\left[ \tfrac{1}{4!}
		\sum_{i_1,\dots,i_4=1}^{d}
		\left[ \tfrac{\partial^{4} u^r}{\partial x_{(i_1)},\dots x_{(i_4)}}
			(x + a \Delta (x)) \prod_{j=1}^{4} \Delta_{(i_4)}(x)\right]  \right] \right|\\
	\le & l^{-2}\frac{d^4}{4!} (2K_0(x)+2K_2(x))K_1(x)
	\end{align*}
	We can deal with the third term similarly to the second term and thus we conclude
	\begin{align*}
		| \E u(x^{x,0}_1) - \E u(\Xtil^{x,0}_1) |
		\leq K(x) l^{-2}
	\end{align*}
\end{proof}

\subsection{One-step approximation}

\lemDtil*
\begin{proof}
	To obtain (i)-(iii), we simply apply Lem.~\ref{app:lem:ito_taylor} with $\psi(z) = \prod_{j=1}^{s} (z_{(i_j)} - x_{(i_j)})$ for $s=1,2,3$ and $i_j\in\{1,\ldots,d\}$ respectively. (iv) is due to \Cref{app:lem:delta_prod_estimate}.
\end{proof}

Next, we estimate the moments of the SVAG iterations below.
\lemD*
\begin{proof}
	Recall $\Delta(x) = - \frac{\eta}{l} \nabla \Loss_{\bgamma}(x)$, where $\Loss_{\bgamma}(x) = \frac{1+\sqrt{2l-1}}{2} \Loss_{\gamma_{1}}(x) + \frac{1-\sqrt{2l-1}}{2}\Loss_{\gamma_{2}}(x)$. Taking expectations, (i) and (ii) are immediate. Note $|\Delta(x)| = O(l^{-0.5})$, (iv) also holds.

\newcommand{\tnabla}{\widetilde{\nabla}}
	Below we show (iii). For convenience, we denote $\sqrt{2l-1}$ by $c$, $\nabla = \nabla \Loss(x)$, $\tnabla_i = \nabla\Loss_{\gamma_i}(x) - \nabla\Loss (x)$, for $i=1,2$ and $\tnabla = \frac{1+c}{2}\tnabla_1+\frac{1-c}{2}\tnabla_2 = \nabla\Loss_\gamma(x) - \nabla\Loss(x)$. We have
	\begin{align*} 
		&\E \frac{l^3}{\eta^3}\Delta(x)^{\otimes 3}
	= \E (\tnabla + \nabla )^{\otimes 3}\\
	= & \E \tnabla ^{\otimes 3} + 3 \E \overline{\tnabla \otimes \tnabla \otimes \nabla} + \nabla^{\otimes 3}\quad (\E \tnabla =0)\\
	= & \E \tnabla ^{\otimes 3} + 3 \E \overline{\Sigma \otimes \nabla} + \nabla^{\otimes 3}\\
	= & \frac{3l-1}{2}\Lambda(x) + 3 \E \overline{\Sigma \otimes \nabla} + \nabla^{\otimes 3},
	\end{align*}
where the last step is because
	\begin{align*} 
		&\E \tnabla ^{\otimes 3}(x)
	= \E (\frac{1+c}{2} \tnabla^1 + \frac{1-c}{2}\tnabla^2 )^{\otimes 3}(x)\\
	= & [((\frac{1+c}{2})^3 + (\frac{1-c}{2})^3)] \E {(\nabla \Loss_{\gamma_1}(x) - \nabla \Loss(x))}^{\otimes 3}\\
	= & [((\frac{1+c}{2})^3 + (\frac{1-c}{2})^3)]  \Lambda(x)\\
	= & \frac{3l-1}{2}  \Lambda(x).
	\end{align*}
\end{proof}

\section{Auxiliary results for the proof of Thm.~\ref{thm:sde_svag}}
\label{app_sec:aux_2}

\begin{lemma}
\label{app:lem:delta_prod_estimate}
	Let $\alpha\ge 1$, there exists a $K\in G$, independent of $l$, such that
	\begin{align*}
		\E \prod_{j=1}^{\alpha}
		\left|
			\Dtil_{(i_j)}
		\right|
		\leq K(x) l^{-\frac{\alpha}{2}}.
	\end{align*}
	where $i_j \in \{ 1,\dots,d \}$ and $C>0$ is independent of $l$.
\end{lemma}
\begin{proof}
	We have
	\begin{align*}
		\E |\Dtil(x)|^{\alpha}
		\leq&
		2^{\alpha-1}
		\E \left|
			\int_{0}^{\frac{\eta}{l}} \nabla \Loss(X^{x,0}_s) ds
		\right|^{\alpha}
		+ 2^{\alpha-1}
		\E \left|
			\int_{0}^{\frac{\eta}{l}} \Sigma^{0.5}(X^{x,0}_s) dW_s
		\right|^{\alpha} \\
		\leq&
		2^{\alpha-1} (\frac{\eta}{l})^{\alpha-1}
		\int_{0}^{\frac{\eta}{l}}
		\E |\nabla \Loss(X^{x,0}_s)|^{\alpha}
		ds
		+ 2^{\alpha-1} 
		\left|
			\int_{0}^{\frac{\eta}{l}} \Sigma^{0.5}(X^{x,0}_s) dW_s
		\right|^{\alpha}
	\end{align*}
	Using Cauchy-Schwarz inequality, It\^{o}'s isometry, we get
	\begin{align*}
		\E \left|
		\int_{0}^{\frac{\eta}{l}} \sigma(X^{x,0}_s) dW_s
		\right|^{\alpha}
		\leq&
		{\left(
			\E \left|
			\int_{0}^{\frac{\eta}{l}} \sigma(X^{x,0}_s) dW_s
			\right|^{2\alpha}
		\right)}^{\nicefrac{1}{2}} \\
		\leq&
		C (\frac{\eta}{l})^{\nicefrac{\alpha-1}{2}}
		{\left(
			\int_{0}^{\frac{\eta}{l}}
			\E | \sigma(X^{x,0}_s) |^{2\alpha }
			ds
		\right)}^{\nicefrac{1}{2}}\\
		= & O(l^{-\nicefrac{\alpha}{2}})
	\end{align*}
	where $C$ depends only on $\alpha$. Now, using the linear growth condition (\ref{thm:sde_svag} (ii)) and the moment estimates in Theorem 19 in \citep{li2019stochastic}, we obtain the result.
\end{proof}

We prove the following It\^{o}-Taylor expansion, which is slightly different from Lemma 28 in \citep{li2019stochastic}.
\begin{lemma}
\label{app:lem:ito_taylor}
	Let $\psi: \R^d \rightarrow \R$ be a sufficiently smooth function. 

	Suppose  that $b, \sigma \in G^3$, and $X^{x,0}_t$ is the solution of the following SDE, with $X^{x,0}_0 = x$. 
	\[ d X_t = b(X_t) dt + \sigma(X_t) dW_t.\]
	Then we have
	\begin{align*}
		\E \psi(X^{x,0}_\eta)
		= \psi(x)
		+ \eta b(x)^\top\nabla \psi(x)
		+ \eta \tr{ \nabla^2\psi \cdot\sigma^2}(x)
		+ \mathcal{O}(\eta^{2}).
	\end{align*}
	
That is,  there exists some function $K\in G$ such that 
	\begin{align*}
		\lvert \E \psi(X^{x,0}_\eta)
		- \psi(x)
		- \eta b(x)^\top\nabla \psi(x)
		- \eta \tr{ \nabla^2\psi \cdot\sigma^2}(x)\rvert \le 
		K(x)\eta^{2}.
	\end{align*}

\end{lemma}
\begin{proof}
	We define operator $A_{1,\epsilon} \psi:=b^\top\nabla \psi, A_{2,\epsilon} \psi := \frac{1}{2}\tr{ \nabla^2\psi \cdot\sigma^2}$.
	
	Using It\^{o}'s formula, we have
	\begin{align*}
		\E\psi(X^{x,0}_\eta)
		=&
		\psi(x)
		+ \int_{0}^{\eta}  \E A_{1,\epsilon} \psi(X^{x,0}_s) ds
		+  \int_{0}^{\eta}\E A_{2,\epsilon} \psi(X^{x,0}_s) ds 
	\end{align*}
	By further application of the above formula to $\E A_{1,\epsilon} \psi$ and $\E A_{2,\epsilon} \psi$, we have
	\begin{align*}
		\E\psi(X^{x,0}_\eta)
		=&
		\psi(x)
		+ \eta A_{1,\epsilon} \psi(x)
		+ \eta A_{2,\epsilon} \psi(x)\\
		&+
		 \int_{0}^{\eta} \int_{0}^{s}
		\E((A_{1,\epsilon}+A_{2,\epsilon})(A_{1,\epsilon}+A_{2,\epsilon})) \psi(X^{x,0}_v) dv ds
	\end{align*}
	Taking expectations of the above, it remains to show that each of the terms is $\mathcal{O}(\eta^2)$. This follows immediately from the assumption that $b, \sigma\in G^3$ and $\psi \in G^4$. Indeed, observe that all the integrands have at most 3 derivatives in $b_0, b_1, \sigma_0$ and 4 derivatives in $\psi$, which by our assumptions all belong to $G$. Thus, the expectation of each integrand is bounded by $ \kappa_1(1 + \sup_{t\in[0,\eta]} \E | X^{x,0}_t |^{2\kappa_2})$ for some $\kappa_1,\kappa_2$, which by Theorem 19 in \citep{li2019stochastic} must be finite. Thus, the expectations of the other integrals are $\mathcal{O}(\eta^2)$ by the polynomial growth assumption and moment estimates in Theorem 19 in \citep{li2019stochastic}.
\end{proof}

We also prove a general moment estimate for the SVAG iterations~\Cref{eq:svag_iter}.
\begin{lemma}
\label{lem:sgd_moment_estimate}
	Let $\{ x_k : k\geq 0 \}$ be the generalized SVAG iterations defined in~\Cref{eq:svag_iter}. Suppose
	\begin{align*}
		| \nabla\Loss_\gamma(x) | \leq L_\gamma ( 1 + | x |), \quad \forall x\in\R^d, \gamma
	\end{align*}
	for some random variable $L_\gamma > 0$ with all moments bounded, i.e., $\E L_\gamma^k <\infty$, for $k\in \N$. Then, for fixed $T>0$ and any $m\geq 1$, $\E | x_k |^{2m} $ exists and is uniformly bounded in $l$ and $k=0,\dots,N \equiv \lfloor lT/\eta \rfloor$.
\end{lemma}

\begin{proof}
Recall that $\nabla \Loss^l_{\bgamma_k}(x) = \frac{1+\sqrt{2l-1}}{2}\nabla \Loss_{\gamma_{k,1}}(x) +  \frac{1-\sqrt{2l-1}}{2}\nabla \Loss_{\gamma_{k,2}}(x)$, thus there exists random variable $L'_\bgamma$ with all moments bounded and $|\nabla \Loss^l_{\bgamma_k}(x)|^2 \le  l (L'_\bgamma)^2 (1+|x|^2)$. We further define $L:= \E L_\gamma$, and thus $|\inp{\E \nabla \Loss^l_{\bgamma_k}(x_k)}{x_k}| \le 2L (1+|x|^2)$.

	For each $k\geq 0$,  we have
	\begin{align*}
		|x_{k+1}|^{2m} \le (1+|x_{k+1}^2|)^m = & \left|1 + |x_k|^2 -2\frac{\eta}{l} \inp{\nabla \Loss^l_{\bgamma_k}(x_k)}{x_k}+ \frac{\eta^2}{l^2}|\nabla \Loss^l_{\bgamma_k}(x_k)| \right|^m	\\
			= & (1+|x_k|)^{2m} - 2m \frac{\eta}{l}\inp{\nabla \Loss^l_{\bgamma_k}(x_k)}{x_k}(1+|x_k|^2)^{m-1} + \frac{1}{l}\cdot O((|x_k|^2+1)^m)
	\end{align*}

	Hence, if we let $a_k:=\E (1+|x_k|^2)^m$, we have
	\begin{align*}
		a_{k+1} =  a_{k+1} -2m \frac{\eta}{l}\inp{\E \nabla \Loss^l_{\bgamma_k}(x_k)}{x_k}(1+|x_k|^2)^{m-1} + \frac{1}{l}\cdot O((|x_k|^2+1)^m)\le 
 (1 +  \frac{C}{l}) a_k 
	\end{align*}
	where $C>0$ are independent of $l$ and $k$, which immediately implies, for all $k=0,\ldots, \lfloor \frac{lT}{\eta}\rfloor$, 
	\begin{align*}
		a_k
		\leq& (1+C/l)^k a_0 \leq (1+C/l)^{lT/\eta}a_0 \le e^{C\frac{CT}{\eta}}a_0.
	\end{align*}
\end{proof}

\section{Omitted proofs in \Cref{sec:failure_mode_1}}\label{appsec:omit_proofs_lsr}
In this section, we provide the missing proofs in \Cref{sec:failure_mode_1}, including \Cref{thm:lsr_condition}, \Cref{thm:lsr_kappa_bound} and the counterpart of \Cref{thm:app_1st_2nd_sde} between $1$st order SDE~\eqref{eq:motivating_sde} and $2$nd order SDE~\eqref{eq:app_2nd_sde_wd}, which is \Cref{thm:2nd_order_sde}. We also provide the derivation of properties for scale invariant functions in \Cref{appsec:sc}.
%

\subsection{Proof of \Cref{thm:eq_sde_sgd}}
\begin{proof}[Proof of \Cref{thm:eq_sde_sgd}]
We will prove the theorem by showing the contrapositive statement: if the equilibriums of~\eqref{eq:sgd_wd} and \eqref{eq:sde_wd} are $C$-close, then $\eta \le \frac{\bNi}{\bGi}(C^2-1)$ and $\frac{1}{C^2-1}\le \frac{\Ni}{\Gi}$.
Following the derivation in \citep{li2020reconciling}, by {\ito}'s lemma and scale invariance of $\Loss_\gamma$:
\begin{align}\label{eq:norm_sde_wd}
\frac{\dd}{\dd t}\E|X_t|^2 = -2\lambda\E |X_t|^2 + \E\Tr[\Sigma(X_t)].	
\end{align}
It can be shown that for SGD~\eqref{eq:sgd_wd}, it holds that
\begin{equation}\label{eq:norm_sgd_wd}	
\begin{split}
	\E|x_{k+1}|^2 -\E |x_{k}|^2=& (1-\eta\lambda)^2\E |x_{k}|^2 + \eta^2\E|\nabla \Loss_{\gamma_k}(x_k)|^2 -\E |x_{k}|^2\\
	\!\!\!\!\!\!= & \eta\lambda(-2\!+\!\eta\lambda)\E |x_{k}|^2\! +\! \eta^2\E |\nabla\Loss(x_k)|^2 +\eta^2\E \Tr[\Sigma(x_k)]
\end{split}
\end{equation}
If both $x_k$ and $X_t$ have reached their equilibriums, both LHS of \eqref{eq:norm_sde_wd} and \eqref{eq:norm_sgd_wd} are $0$, and therefore 
\begin{align}
	(2-\eta\lambda)\lambda\Ri &= \eta \Gi + \eta\Ni, \label{eq:RGN_sgd}\\
	2\lambda\bRi &= \phantom{\eta \bGi+\eta }  \bNi. \label{eq:RGN_sde}
\end{align}
Combining \eqref{eq:RGN_sgd}, \eqref{eq:RGN_sde}, and \eqref{eq:C_close}, we have 
\[\eta\Gi + \eta\Ni \le 2\lambda \Ri \le 2\lambda C\bRi= C\bNi.\]
Applying \eqref{eq:C_close} again, we have 
\( \eta\bGi+ \bNi \le C\eta(\Gi + \Ni) \le  C^2\bNi \le C^3\eta \Ni.\)
\end{proof}

\subsection{Proof of \Cref{thm:lsr_kappa_bound}}
\begin{proof}[Proof of \Cref{thm:lsr_kappa_bound}]
Suppose $(C,\kappa)$-LSI hold, similar to \Cref{eq:RGN_sgd}, we have
\begin{align}
    (2-\kappa\eta\lambda)\lambda\Ri^{\kappa B,\kappa \eta} &= \kappa\eta (\Gi^{\kappa B,\kappa \eta}+\Ni^{\kappa B,\kappa \eta}), \label{eq:lb_RGN_sgd}\\
    (2-\eta\lambda)\lambda\Ri^{B,\eta} &= \eta (\Gi^{B,\eta} + \Ni^{B,\eta}), \label{eq:sb_RGN_sgd}
\end{align}

Thus combining  \eqref{eq:lb_RGN_sgd} ,\eqref{eq:sb_RGN_sgd} and \eqref{eq:C_close}, we have 
\[\kappa(\Gi^{\kappa B,\kappa \eta} + \Ni^{\kappa B,\kappa \eta})= (2-\kappa\eta\lambda)\frac{ \lambda}{\eta} \Ri^{\kappa B,\kappa \eta} \le (2-\eta\lambda)\frac{\lambda}{\eta} C\Ri^{B,\eta}= C(\Ni^{B,\eta} + \Gi^{B,\eta}).\]

Applying \eqref{eq:C_kappa_close} again, we have 
\[ \kappa \Gi^{ B} + \Ni^{ B} \le C\kappa (\Gi^{\kappa B,\kappa \eta} +\Ni^{\kappa B,\kappa \eta}) \le  C^2(\Ni^{B,\eta} + \Gi^{B,\eta}) .\]

Therefore we conclude that $\kappa \le C^2(1+\frac{\Ni^{B,\eta}}{\Gi^{B,\eta}})$.

\end{proof}

\subsection{Proof of \Cref{thm:lsr_condition}}
\begin{proof}
Suppose $(C,\kappa)$-LSI hold, by \eqref{eq:C_close}, we have 
\[CG^{\kappa B,\kappa \eta}_\infty + C\kappa N^{\kappa B,\kappa \eta}_\infty	
 \ge G^{ B, \eta}_\infty + N^{ B, \eta}_\infty .\]
 
By \eqref{eq:lb_RGN_sgd} ,\eqref{eq:sb_RGN_sgd} and \eqref{eq:C_close}, we have 
\begin{equation}
 G^{ B, \eta}_\infty + N^{ B, \eta}_\infty = 2\lambda \frac{B}{\eta} R^{ B, \eta}_\infty \ge  \frac{2}{C}\lambda \frac{B}{\eta} R^{\kappa B,\kappa \eta}_\infty =
\frac{\kappa}{C}(G^{\kappa B,\kappa \eta}_\infty + N^{\kappa B,\kappa \eta}_\infty).
\end{equation}

Rearranging things, we have 
\[  N^{\kappa B,\kappa \eta}_\infty \ge \frac{\kappa -C^2}{\kappa(C^2-1)} G^{\kappa B,\kappa \eta}_\infty\ge ( (1-\frac{1}{\kappa})\frac{1}{C^2-1} -\frac{1}{\kappa})G^{\kappa B,\kappa \eta}_\infty.\]
\end{proof}

%
%

\subsection{Necessary condition for $C$-closeness between $1$st order and $2$nd order SDE approximation}

In this section we will present a necessary condition for $C$-closeness between $1$st order and $2$nd order SDE approximation, similar to that betweeen $1$st order approximation and SGD. The key observation is that the missing second order term $\eta\Gi$ in $1$st order SDE, also appears in the $2$nd order SDE, as it does for SGD. Thus we can basically apply the same analysis and show the similar conclusion~(\Cref{thm:app_1st_2nd_sde}).  

Below we first recap the notion of $1$st and $2$nd order SDE approximation with weight decay (i.e., $\ell_2$ regularization). We first define $\Loss'_{\gamma}(X) = \Loss_{\gamma}(X) + \frac{\lambda}{2} |X|^2$ and the SGD dynamics~\eqref{eq:app_sgd_wd} can be written by 
\begin{align}\label{eq:app_sgd_wd}
x_{k+1} = x_k - \eta\nabla \Loss'_\gamma(x_k)= x_k - \eta \nabla \big(\Loss_{\gamma_k}(X_t) + \frac{\lambda}{2} |X_t|^2\big)
\end{align}

Below we recap the $1$st and $2$nd order SDE approximation:

\begin{itemize}
\item $1$st order SDE approximation (with $\overline{\Sigma} = \eta\Sigma$):	
	\begin{align}\label{eq:app_1st_sde_wd}
\dd X_t = - \nabla \big(\Loss(X_t) + \frac{\lambda}{2} |X_t|^2\big) \dd t +  {(\eta\Sigma)^{1/2}(X_t)} \dd W_t
\end{align}

\item $2$nd order SDE approximation:	
	\begin{align}\label{eq:app_2nd_sde_wd}
\dd X_t = - \nabla \big( \Loss'(X_t) + \frac{\eta}{4} |\nabla \Loss'(X_t)|^2 \big)\dd t +  {(\eta\Sigma)^{1/2}(X_t)} \dd W_t
\end{align}
\end{itemize}

\begin{theorem}[Theorem 9 in \cite{li2019stochastic}]
\label{thm:2nd_order_sde}
\eqref{eq:app_2nd_sde_wd} is an order-2 weak approximation of SGD~\eqref{eq:sgd_iter}:
\end{theorem}

We first prove a useful lemma.
\begin{lemma}\label{lem:2nd_sde_hessian}
Suppose $\Loss$ is scale invariant, then for any $X\in\R^d$, $X\neq 0$, 
\[X^\top \nabla^2\Loss(X) \nabla\Loss(X) = \frac{1}{2}X^\top \nabla(\norm{\nabla\Loss}_2^2) =-  \norm{\nabla\Loss(X)}_2^2.\]
\end{lemma}

\begin{proof} 
By chain rule, we have
\begin{align*}
 &X^\top \nabla(\norm{\nabla\Loss}_2^2) \\
 = &\lim_{t\to 0}\frac{\norm{\nabla\Loss((1+t)X)}_2^2 - \norm{\nabla\Loss(X)}_2^2}{t} \\
 = &\lim_{t\to 0 }\frac{(1+t)^{-2}-1}{t}\norm{\nabla\Loss(X)}_2^2 \quad (\text{by scale invariance})\\
 = &-2\norm{\nabla\Loss(X)}_2^2
\end{align*}
\end{proof}

\begin{definition}[$C$-closeness]\label{def:app_lsi}
We use $\bRi':= \lim\limits_{t\to \infty} E|X_t|^2, \bGi':= \lim\limits_{t\to \infty}\E |\nabla \Loss(X_t)|^2, \bNi':= \lim\limits_{t\to \infty} \E[\Tr[\overline{\Sigma}(X_t)] = \lim\limits_{t\to \infty} \E[\Tr[\eta\Sigma(X_t)]$ 
to denote the limiting squared norm, gradient norm and trace of covariance for SDE~\eqref{eq:app_2nd_sde_wd}. (We assume both$X_t$ converge to their equilibrium so the limits exist). 
We say the two equilibriums of $1$st order SDE approximation~\eqref{eq:app_1st_sde_wd} and $2$nd order SDE approximation~\eqref{eq:2nd-sgd-sde-wd} are \emph{$C$-close} to each other iff 
\begin{align}\label{eq:app_C_close}
	\frac{1}{C}\le \frac{\bRi}{\bRi'},\frac{\bGi}{\bGi'},\frac{\bNi}{\bNi'}\le C.
\end{align}
\end{definition}

The following theorem is an analog of \Cref{thm:eq_sde_sgd}.
\begin{theorem}\label{thm:app_1st_2nd_sde}
If the equilibriums of~\eqref{eq:app_1st_sde_wd} and \eqref{eq:app_2nd_sde_wd} exist and are $C$-close for some $C>0$, then
	\[\eta \le \frac{\bNi}{\bGi}\big(C^2(1+\frac{\eta\lambda}{2})-1 \big)\approx \frac{\bNi}{\bGi}\big(C^2-1 \big),\]
\end{theorem}

where $\lambda$ is usually of scale $10^{-4}$ in practice and thus can be omitted when calculating upper bound.
\begin{proof}[Proof]
Since $\Loss$ is scale-invariant, so $\nabla\Loss(X)^\top X =0$, which implies $|\nabla\Loss'(X)|^2 = |\nabla\Loss(X)|^2 + \lambda^2|X|^2$. Plug in $\Loss'$, we have

\begin{equation} \label{eq:2nd-sgd-sde-wd}
	\dd X_{t} = -  \left(\nabla \Loss(X_t) + \frac{\eta}{2} \nabla^2\Loss(X_t)\nabla\Loss(X_t)\right)  \dd t+ (\eta\Sigma)^{1/2}(X_t) \dd W_t - \lambda(1+\frac{\eta\lambda}{2})X_t \dd t.
\end{equation}	

Applying {\ito}' lemma, we have 
\begin{equation}\label{eq:app_ito_lemma}
\begin{split}
\dd |X_t|^2 
= & 2\inp{X_t}{\dd X_t} + \inp{\dd X_t}{\dd X_t}\\
= & -2\lambda(1+\frac{\eta\lambda}{2})|X_t|^2 \dd t  -\eta\inp{X_t}{\nabla^2\Loss(X_t)\nabla\Loss(X_t)}\dd t
+  \cancel{$2\inp{X_t}{\Sigma^{1/2}(X_t)\dd W_t}$} - \cancel{$2 \inp{X_t}{\nabla\Loss(X_t)}\dd t$}\\
+ & \Tr[\Sigma(X_t)]\dd t\\
= & -2\lambda(1+\frac{\eta\lambda}{2})|X_t|^2 \dd t  -\eta\inp{X_t}{\nabla^2\Loss(X_t)\nabla\Loss(X_t)}\dd t
+  \Tr[\eta\Sigma(X_t)]\dd t \quad \text{(by \Cref{cor:sc})}\\
= & -2\lambda(1+\frac{\eta\lambda}{2})|X_t|^2 \dd t  + \big(\eta\Tr[\Sigma(X_t)] + \eta|\nabla \Loss(X_t)|^2\big)\dd t \quad\quad \text{(by \Cref{lem:2nd_sde_hessian})}
\end{split}
\end{equation}

Thus 
\begin{equation}
\frac{\dd \E[|X_t|^2]}{\dd t} = -2 \lambda(1+\frac{\eta\lambda}{2})\E[|X_t|^2] +  \E[\eta\Tr[\Sigma(X_t)] + \eta|\nabla\Loss(X_t)|^2].	
\end{equation}

Suppose $X_t$ is samples from the equilibrium, we have $\frac{\dd \E[|X_t|^2]}{\dd t}=0$ and 
\begin{align}\label{eq:app_RGN_2nd_sde}
	(2+\eta\lambda)\lambda\bRi' &= \eta \bGi' + \bNi'. 
\end{align}

If we compare \Cref{eq:app_RGN_2nd_sde} to \Cref{eq:RGN_sgd,eq:RGN_sde} (we recap them below), it's quite clear $2$nd order is much closer to SGD in terms of the relationship between $\Ri$, $\Gi$ and $\Ni$. Thus  $1$st and $2$nd order SDE approximation won't be $C$-close if $\frac{\bNi}{\bGi}$ is larger than some constant for the exact same reason that $1$st SDE is not $C$-close to SGD. 
\begin{align}
	(2-\eta\lambda)\lambda\Ri &= \eta \Gi + \eta\Ni, \tag{\ref{eq:RGN_sgd}}\\
	2\lambda\bRi &= \phantom{\eta \bGi} + \bNi. \tag{\ref{eq:RGN_sde}}
\end{align}

In detail, by combining \eqref{eq:app_RGN_2nd_sde}, \eqref{eq:RGN_sde} and \eqref{eq:app_C_close}, we have 
\[\eta\bGi' + \bNi' = (2+\eta\lambda) \lambda \bRi' \le (2+\eta\lambda)\lambda C\bRi= (1+\frac{\eta\lambda}{2})C\bNi.\]
Applying \eqref{eq:app_C_close} again, we have 
\[ \eta\bGi+ \bNi \le C(\eta\bGi' + \bNi') \le  C^2(1+\frac{\eta\lambda}{2})\bNi\le C^3(1+\frac{\eta\lambda}{2})\bNi',\]
which imples $\eta \le \big(C^2(1+\frac{\eta\lambda}{2})-1\big)\min\{\frac{\bNi}{\bGi},\frac{\bNi'}{\bGi'}\}$.
\end{proof}
\begin{remark}
	\cite{kunin2020neural} derived a similar equation to \Cref{eq:app_ito_lemma} in Appendix F of their paper.
\end{remark}


\subsection{Properties of Scale Invariance Function}\label{appsec:sc}
These properties are proved in \cite{arora2018theoretical}. We include them here for self-containedness.

\begin{definition}[Scale Invariance]
	We say $\Loss:\R^d\to R$ is \emph{scale invariant} iff $\forall x\in \R^d, x\neq 0$ and $\forall c>0$, it holds that
	\[ \Loss(cx) = \Loss(x).\]
\end{definition}

We have the following two properties.
\begin{lemma}\label{lem:sc}
	If $\Loss$ is scale invariant, then $\forall x\in\R^d/\{0\}$, we have 
\begin{enumerate}
\item $\inp{x}{\nabla \Loss(x)}=0$.
\item $\forall c>0$, $c\nabla \Loss(cx) = \nabla \Loss(x)$.
\end{enumerate}
\end{lemma}

\begin{proof}
	For (1), by chain rule, we have $\inp{x}{\nabla \Loss(x)} = \lim_{t\to 0}\frac{\Loss((1+t)x)-\Loss(x)}{ t} =0$.
	
	For (2), for any $v\in\R^d$, again by chain rule, we have 
	\[\inp{v}{c\nabla \Loss(cx)}  =\inp{cv}{\nabla \Loss(cx)}= \lim_{t\to 0}\frac{\Loss(cx+cvt) - \Loss(cx)}{t} =
	\lim_{t\to 0}\frac{\Loss(x+vt) - \Loss(x)}{t}  = \inp{v}{\nabla \Loss(x)}.\]
\end{proof}

Suppose $\Loss_\gamma$ is a random loss and is scale invariant for every $\gamma$, and we use $\nabla \Loss(x)$ and $\Sigma(x)$ to denote the expectation and covariance of gradient, then for any $x\in\R^d/\{0\}$, we have the following corollary:

\begin{corollary}\label{cor:sc}
	$\inp{x}{\nabla \Loss(x)} =0$, $x^\top \Sigma(x)x =0$.
\end{corollary}


\newpage
\section{Experiments}\label{sec:app_exp}

We use the models from Github Repository: \url{https://github.com/bearpaw/pytorch-classification}. For VGG and PreResNet, unless noted otherwise, we modified the model following Appendix C of \citep{li2020exp} so that the network is scale invariant, e.g., fixing the last layer. Such modification doesn’t lead to change in performance, as shown in \cite{hoffer2018fix}.
We use Weights \& Biases to manage our experiments \citep{wandb}. 


\subsection{Further Verification of SVAG}\label{subsec:app_svag_exp}
We verify that SVAG converges for different architectures (including ones without normalization), learning rate schedules, and datasets. We further conclude that for most the standard settings we consider (excluding the use of large batch size in~\Cref{fig:svag_accs,fig:app_svag_svhn} and GroupNorm on CIFAR-100 in~\Cref{fig:app_svag_c100}), SVAG with large $l$ achieves similar performance to SGD, i.e. SVAG with $l=1$.

\subsubsection{SVAG and Sampling Methods}
\begin{figure}[!htbp]
	\centering
	\includegraphics[width=\linewidth]{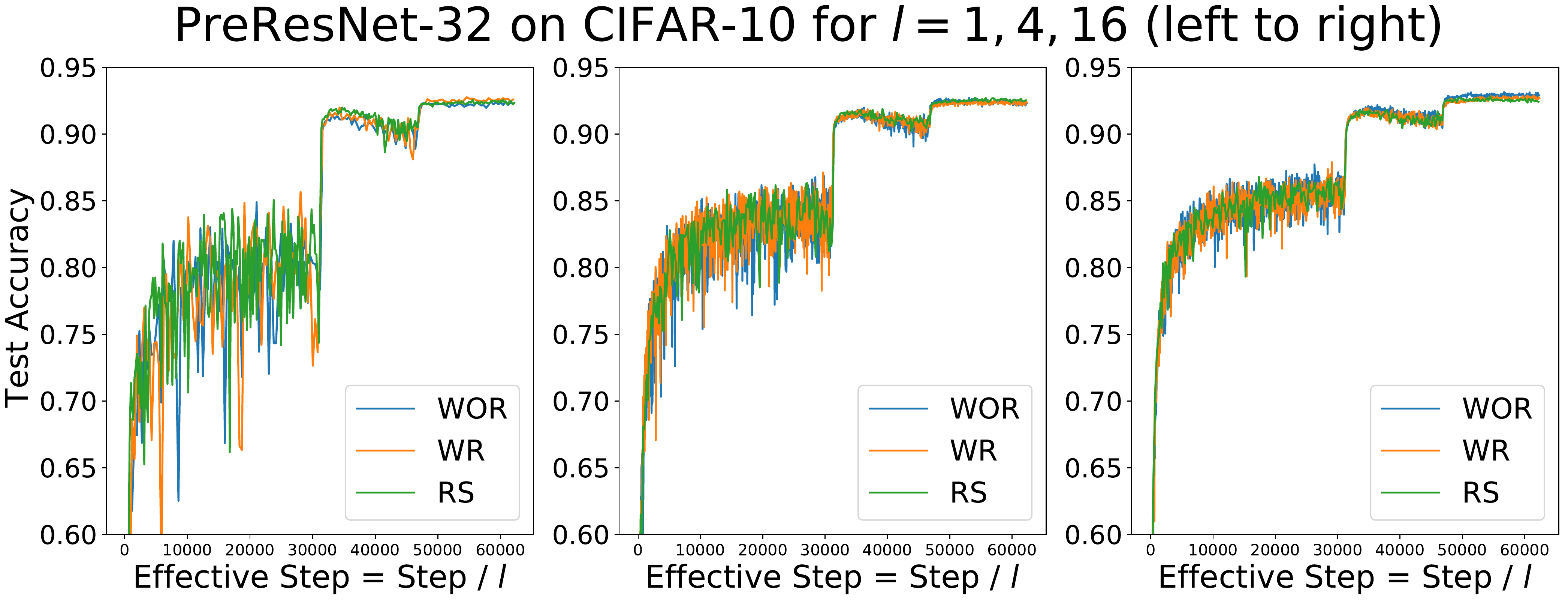}
	\caption{\small 
		Different sampling methods have little impact on the performance of SVAG (SGD). We compare random shuffling (RS), which is the default implementation in PyTorch, to with replacement (WR) and without replacement (WOR), which \Cref{thm:sde_svag} applies to. We train for 320 epochs with initial LR $\eta=0.8$ with 2 LR decays by a factor of 0.1 at epochs 160 and 240, and we use weight decay with $\lambda = 5\mathrm{e}{-4}$ and batch size $B=128$. Since SVAG takes $l$ smaller steps to simulate the continuous dynamics in $\eta$ time, we plot accuracy against ``effective steps'' defined as $\frac{\#\text{steps}}{ l}$. 
	}	
	\label{fig:app_svag_sampling}
\end{figure}

\Cref{thm:sde_svag} only holds if each step of SGD is a Markov process, which is in part determined by how each example is sampled from the dataset.
We describe three common ways that examples can be sampled from the dataset during training.
\begin{enumerate}
	\item \emph{Random shuffling (RS)}: RS is standard practice in experiments and is the default implementation in PyTorch. A random shuffled order of datapoints is fixed at the start of each epoch, and each sample is drawn in this order. SGD with this sampling scheme can be viewed as a Markov process per epoch, although not per step. We use this method in our experiments, but our theory for SVAG (\Cref{sec:svag}) does not cover this sampling method.
	\item \emph{Without replacement (WOR)}: WOR requires drawing each sample i.i.d. from the dataset without replacing previously drawn ones. SGD with this sampling scheme can be viewed as a Markov process per step, so our theory for SVAG (\Cref{sec:svag}) does cover this case.
	\item \emph{With replacement (WR)}: WR requires drawing each sample i.i.d. from the dataset with replacement. SGD with this sampling scheme can be viewed as a Markov process per step, so our theory for SVAG (\Cref{sec:svag}) does cover this case.
\end{enumerate}
In \Cref{fig:app_svag_sampling}, we observe that SVAG (including SGD) behaves similarly when using all three of these sampling methods. 
Therefore, although our theory does not directly apply to the commonly used RS scheme, we can heuristically apply \Cref{thm:sde_svag} to understand its behavior.

We furthermore note that our findings do not match the conclusion in \cite{smith2021origin} that SGD with RS has a different implicit bias compared to WOR and WR. We suggest two possible reasons for this discrepancy: 
(1) Their result holds when $\eta$ is smaller than an unmeasurable constant, so it may be the case that their results do not apply to the constant LR regime we use SVAG in. (2) Their result concerns behavior after a single epoch and our experiments run for hundreds of epochs.

\subsubsection{Further Verification of SVAG on more architectures on CIFAR-10}
For all CIFAR-10 experiments in this subsection, there are 320 epochs with initial LR $\eta=0.8$ and 2 LR decays by a factor of 0.1 at epochs 160 and 240 and we use weight decay with $\lambda = 5\mathrm{e}{-4}$ and batch size $B=128$.
We also use the standard data augmentation for CIFAR-10: taking a random $32\times 32$ crop after padding with 4 pixels and randomly performing a horizontal flip.

\begin{figure}[!htbp]
    \centering
    \includegraphics[width=\linewidth]{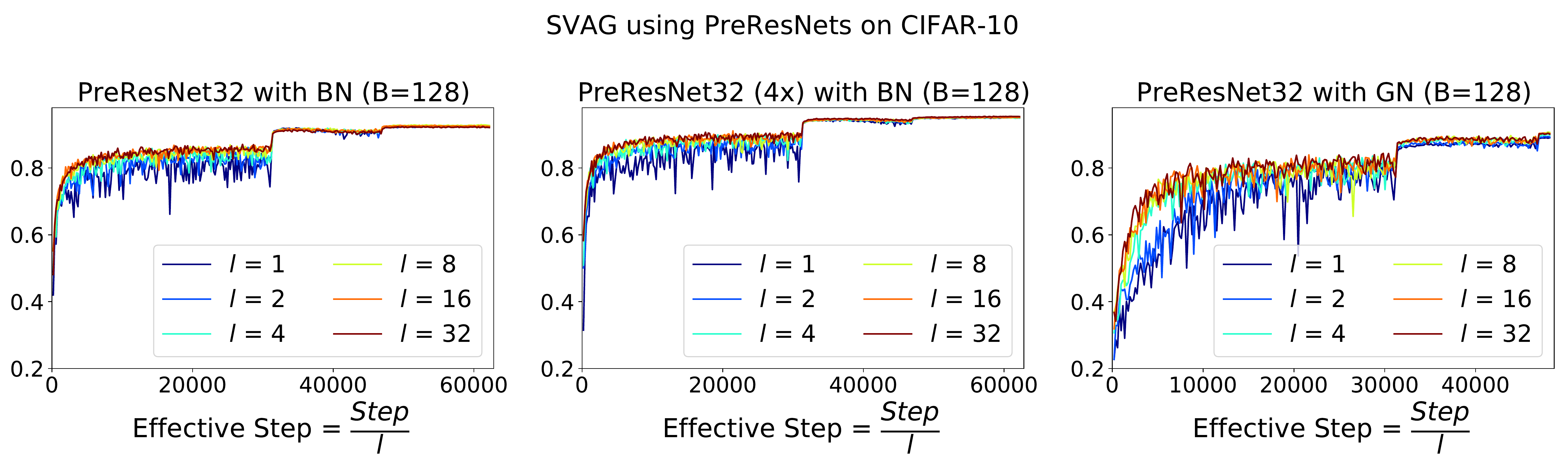}
    \caption{\small Validation accuracy for PreResNet32 with BatchNorm (left), PreResNet32 (4x) with BatchNorm (middle) and PreResNet32 with GroupNorm (right) during training on CIFAR-10. SVAG converges and closely follows the SGD trajectory in all three cases.   
	Since SVAG takes $l$ smaller steps to simulate the continuous dynamics in $\eta$ time, we plot accuracy against ``effective steps'' defined as $\frac{\#\text{steps}}{ l}$. 
    }
    \label{fig:app_svag_baselines}
\end{figure}

In~\Cref{fig:app_svag_baselines}, we demonstrate that SVAG converges and closely follows SGD for PreResNet32 with BatchNorm (left), PreResNet32 (4x) with BatchNorm (middle) and PreResNet32 with GroupNorm (right).

\begin{figure}[!htbp]
    \centering
    \includegraphics[width=\linewidth]{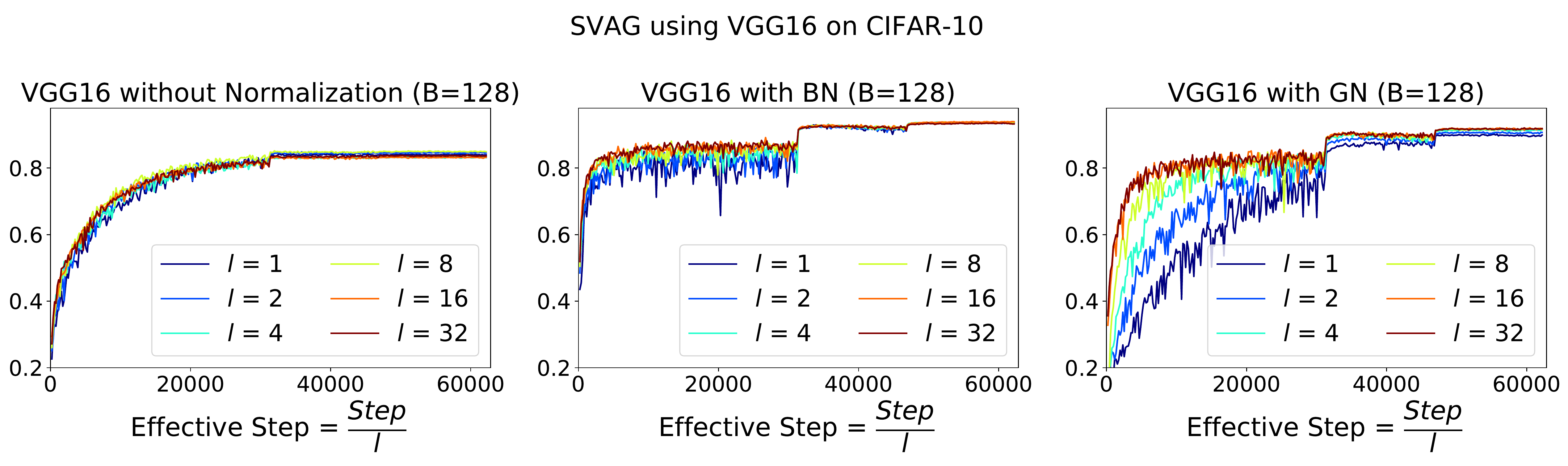}
    \caption{\small Validation accuracy for VGG16 without Normalization (left), VGG16 with BatchNorm (middle) and VGG16 with GroupNorm (right) during training on CIFAR-10. SVAG converges and closely follows the SGD trajectory in all three cases.       	Since SVAG takes $l$ smaller steps to simulate the continuous dynamics in $\eta$ time, we plot accuracy against ``effective steps'' defined as $\frac{\#\text{steps}}{ l}$. 
    }
    \label{fig:app_svag_otherarch}
\end{figure}

In~\Cref{fig:app_svag_otherarch}, we demonstrate that SVAG converges and closely follows SGD for VGG16 without Normalization (left), VGG16 with BatchNorm (middle) and VGG16 with GroupNorm (right).

\subsubsection{Further Verification of SVAG on more complex LR schedules}
\begin{figure}[!htbp]
    \centering
    \includegraphics[width=0.6\linewidth]{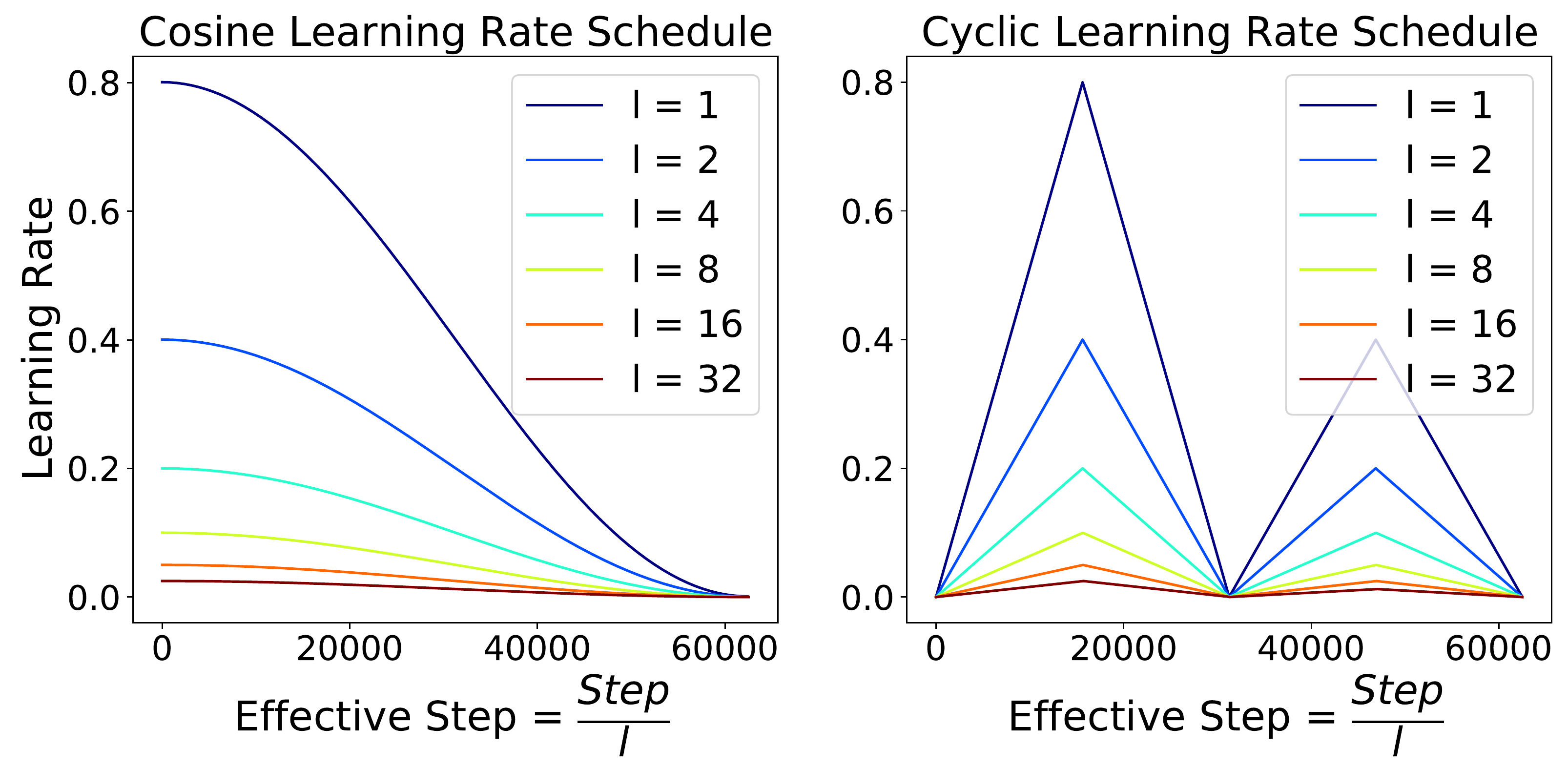}
    \caption{\small Cosine (left, \cite{loshchilov2016sgdr}) and cyclic (right, \cite{smith2017cyclical}) learning rate schedules for different SVAG configurations, plotted against ``effective steps'' defined as $\frac{\#\text{steps}}{ l}$. 
    }
    \label{fig:app_exotic_lr}
\end{figure}

\begin{figure}[!htbp]
    \centering
    \includegraphics[width=0.9\linewidth]{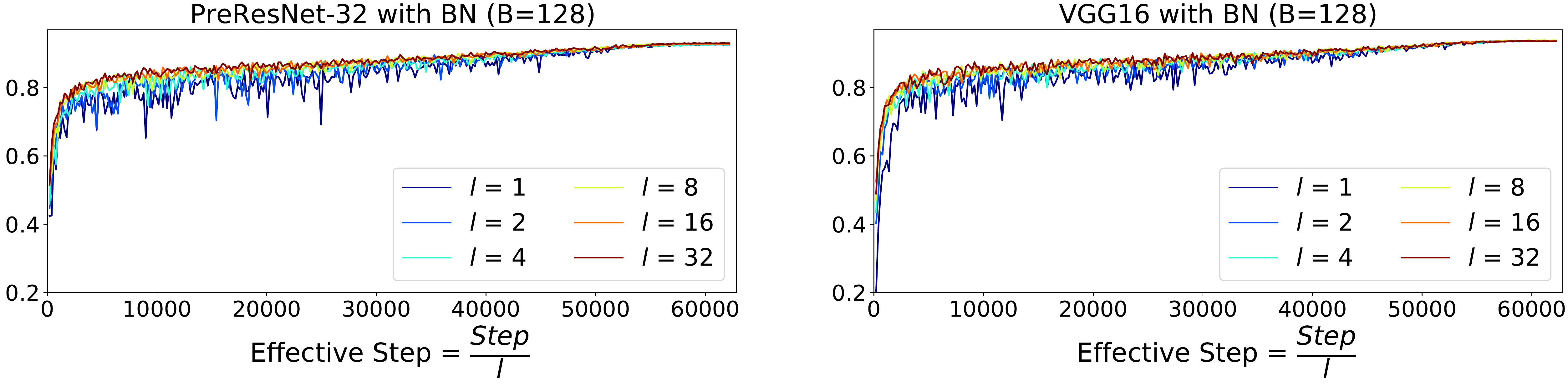}
    \caption{\small Validation accuracy for PreResNet-32 with $B=128$ (left) and VGG16 with $B=128$ (right) during training on CIFAR-10. We use the triangle LR schedule proposed in \cite{smith2017cyclical} with 80 epochs of increase to LR 0.8, 80 epochs of decay to 0, 80 epochs of increase to 0.4 and 80 epochs of decay to 0.
    The LR schedule is visualized in \Cref{fig:app_exotic_lr}.
   	Since SVAG takes $l$ smaller steps to simulate the continuous dynamics in $\eta$ time, we plot accuracy against ``effective steps'' defined as $\frac{\#\text{steps}}{ l}$. 
    }
    \label{fig:app_svag_triangle}
\end{figure}

We verify that SVAG converges and closely follows the SGD trajectory for networks trained with more complex learning rate schedules. 
In~\Cref{fig:app_svag_triangle}, we use the triangle (i.e., cyclical) learning rate schedule proposed in \cite{smith2017cyclical}, visualized in \Cref{fig:app_exotic_lr}.
We implement the schedule over 320 epochs of training: we increase the initial learning rate $0.001$ linearly to $0.8$ over $80$ epochs, decay the LR to $0.001$ over the next $80$ epochs, increase the LR to $0.4$ over $80$ epochs, and decay the LR to $0.001$ over the remaining 80 epochs.
As seen in \Cref{fig:app_svag_triangle}, SVAG converges to the SGD trajectory in this setting.

\begin{figure}[!htbp]
    \centering
    \includegraphics[width=0.9\linewidth]{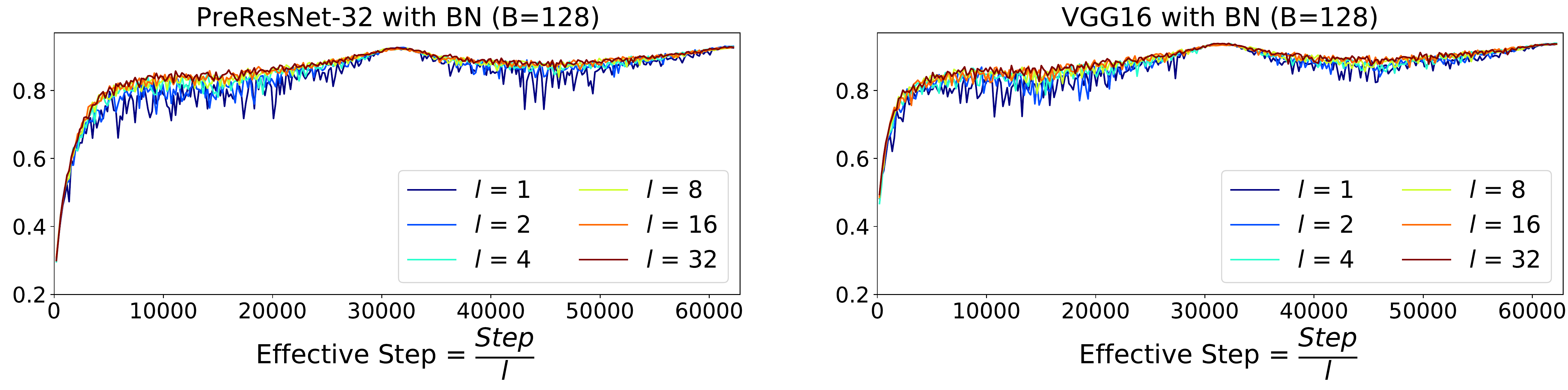}
    \caption{\small Validation accuracy for PreResNet-32 with $B=128$ (left) and VGG16 with $B=128$ (right) during training on CIFAR-10. We use the cosine LR schedule proposed in \cite{loshchilov2016sgdr} starting with LR 0.8 and following the cosine curve until the LR becomes infinitesimally small.
    The LR schedule is visualized in \Cref{fig:app_exotic_lr}.
       	Since SVAG takes $l$ smaller steps to simulate the continuous dynamics in $\eta$ time, we plot accuracy against ``effective steps'' defined as $\frac{\#\text{steps}}{ l}$. 
    }
    \label{fig:app_svag_cosine}
\end{figure}

We further test SVAG on the cosine learning rate schedule proposed in \cite{loshchilov2016sgdr} with $\eta_{\text{max}}=0.8 $ and $\eta_{\text{min}}=0.001$ with total training budgets of $160$ epochs. We visualize the schedule in \Cref{fig:app_exotic_lr}. In~\Cref{fig:app_svag_cosine}, we see that SVAG converges and closely follows the SGD trajectory, suggesting the SDE~\eqref{eq:motivating_sde} can model SGD trajectories with complex learning rate schedules as well.

\subsubsection{Further Verification of SVAG on more datasets (CIFAR-100 and SVHN)}
\begin{figure}[!htbp]
    \centering
    \includegraphics[width=\linewidth]{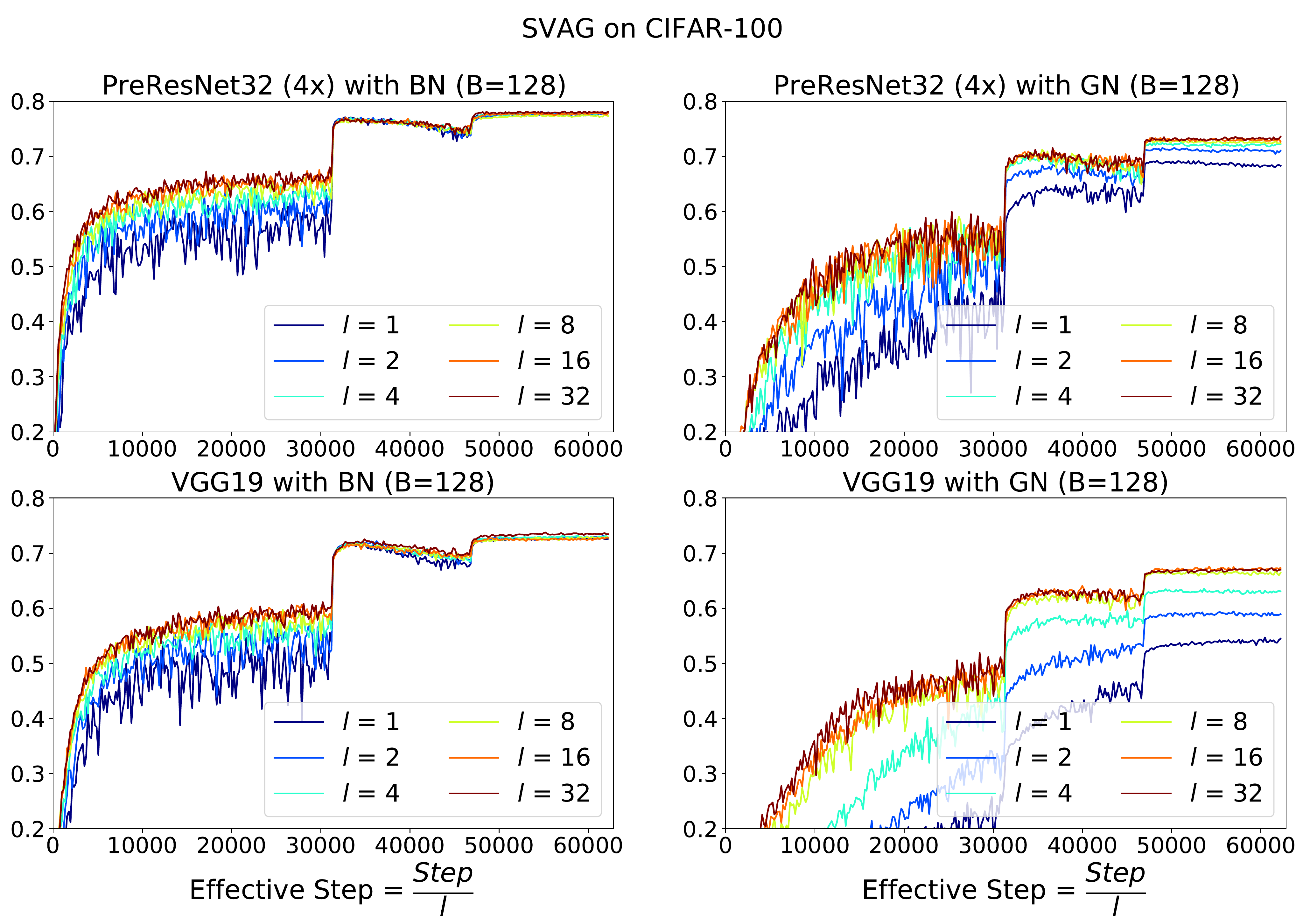}
    \caption{\small Validation accuracy for wider PreResNet-32 with $B=128$ using BN (top left) and GN (top right) and for VGG-19 with $B=128$ using BN (bottom left) and GN (bottom right) trained on CIFAR-100.
    We train for 320 epochs and decay the LR by 0.1 at epochs 160 and 240.
       	Since SVAG takes $l$ smaller steps to simulate the continuous dynamics in $\eta$ time, we plot accuracy against ``effective steps'' defined as $\frac{\#\text{steps}}{ l}$. Interestingly, we found the performance of BatchNorm out performs GroupNorm and the performance of the latter gets improved when using larger $l$ for SVAG.
    }
    \label{fig:app_svag_c100}
\end{figure}

\begin{figure}[!htbp]
    \centering
    \includegraphics[width=\linewidth]{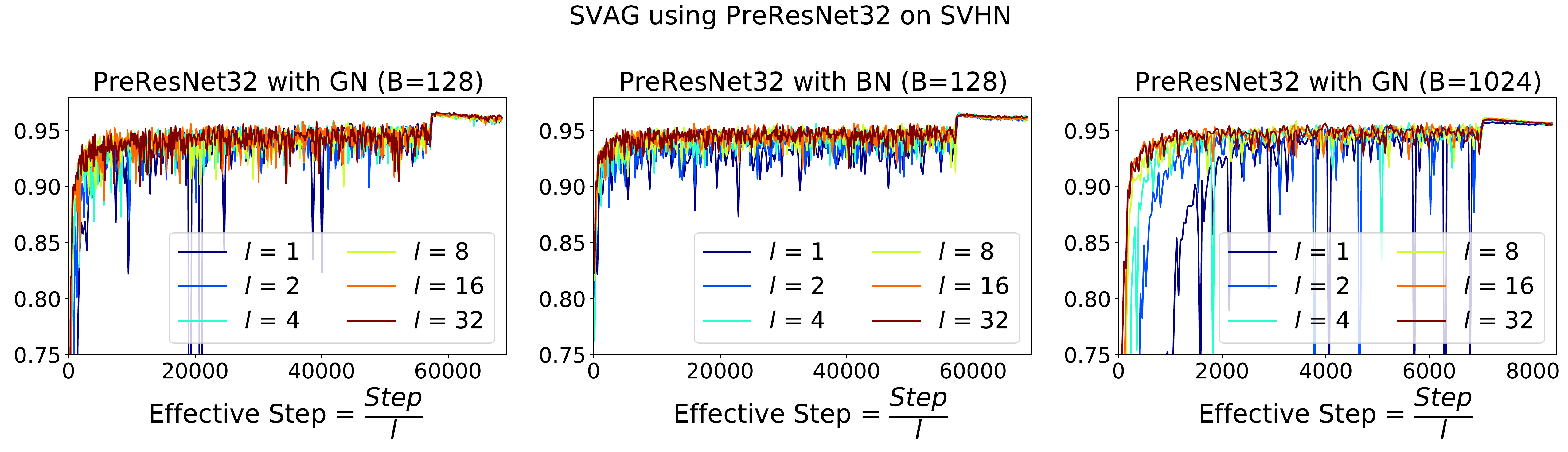}
    \caption{\small Validation accuracy for PreResNet-32 with $B=128$ using GN (left) and BN (center) and with $B=1024$ using GN trained on the SVHN dataset~\citep{netzer2011reading}. 
    We train for 240 epochs and decay the LR by 0.1 at epoch 200.  
       	Since SVAG takes $l$ smaller steps to simulate the continuous dynamics in $\eta$ time, we plot accuracy against ``effective steps'' defined as $\frac{\#\text{steps}}{ l}$. 
    }
    \label{fig:app_svag_svhn}
\end{figure}

We also verify that SVAG converges on the CIFAR-100 dataset. We set the learning rate to be $0.8$ and decay it by a factor of $0.1$ at epochs $160$ and $240$ with a total budget of $320$ epochs. 
We use weight decay with $\lambda = 5\mathrm{e}{-4}$. 
We use the standard data augmentation for CIFAR-100: randomly taking a $32\times 32$ crop from the image after padding with 4 pixels and randomly horizontally flipping the result.
We observe that SVAG converges for computationally tractable value of $l$ in ~\Cref{fig:app_svag_c100}, but  for both GN architectures, the SDE fails to approximate SGD training. 

We also verify SVAG on the Street View House Numbers (SVHN) dataset~\citep{netzer2011reading}. We set the learning rate to $0.8$ for batch size $128$ and scale it according to LSR (\Cref{def:lsr}) for large batch training.
We train for $240$ epochs and decay the learning rate by a factor of $0.1$ once at epoch $200$. We use weight decay with $\lambda = 5\mathrm{e}{-4}$. 

\newpage
\subsection{Further Verification of Necessary Condition for LSR}\label{subsec:app_lsr_exps}
\begin{figure}[!htbp]
    \centering
    \includegraphics[width=0.9\linewidth]{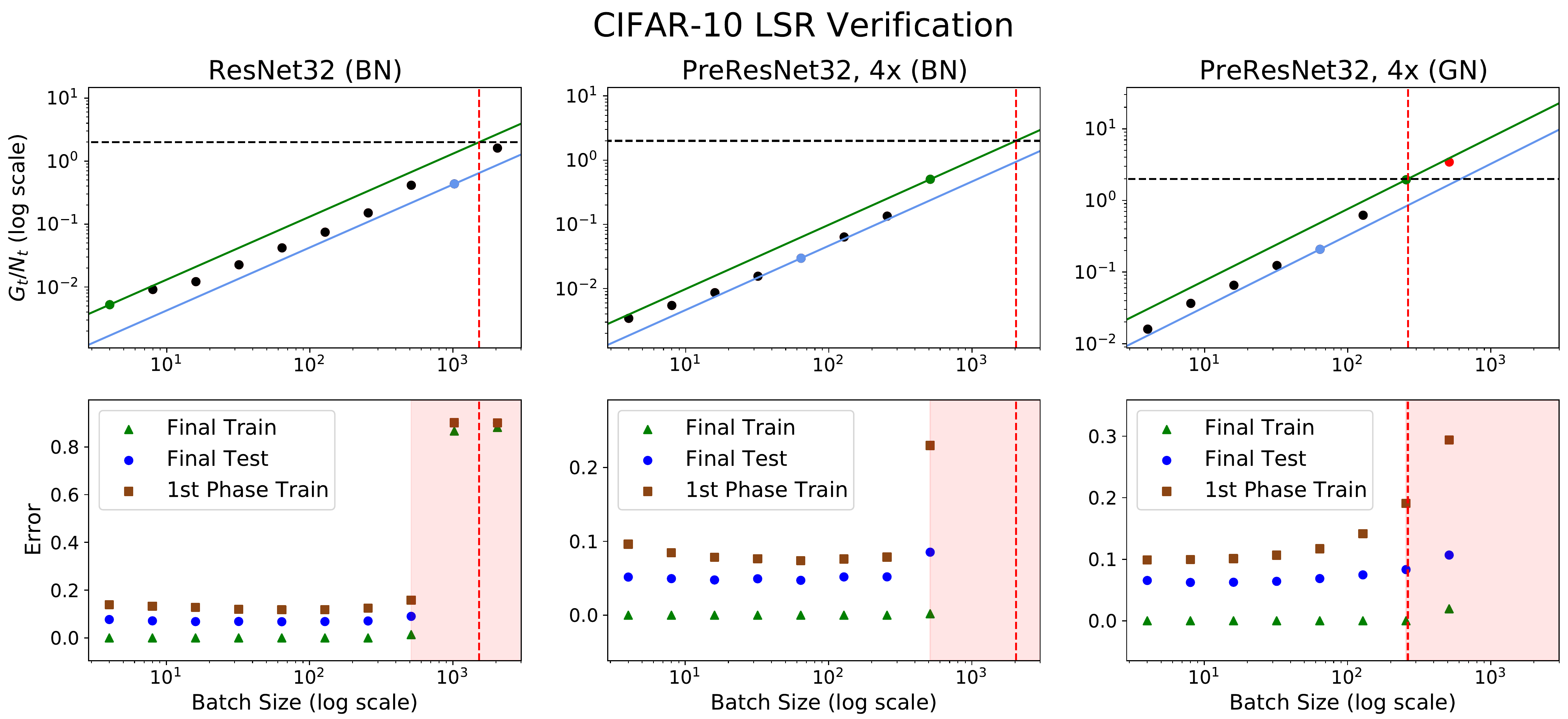}
    \caption{\small Further verification for our theory on predicting the failure of Linear Scaling Rule. We test if the condition applies to different architectures trained on CIFAR-10. All three settings use the same LR schedule, LR$=0.8$ initially and is decayed by $0.1$ at epoch $250$ with $300$ epochs total budget.   We measure $G_t$ and $N_t$ by averaging their values over the last 50 epochs of the first phase (i.e., from epoch 200 to 250).}
    \label{fig:app_lsr_c10}
\end{figure}

\begin{figure}[!htbp]
    \centering
    \includegraphics[width=\linewidth]{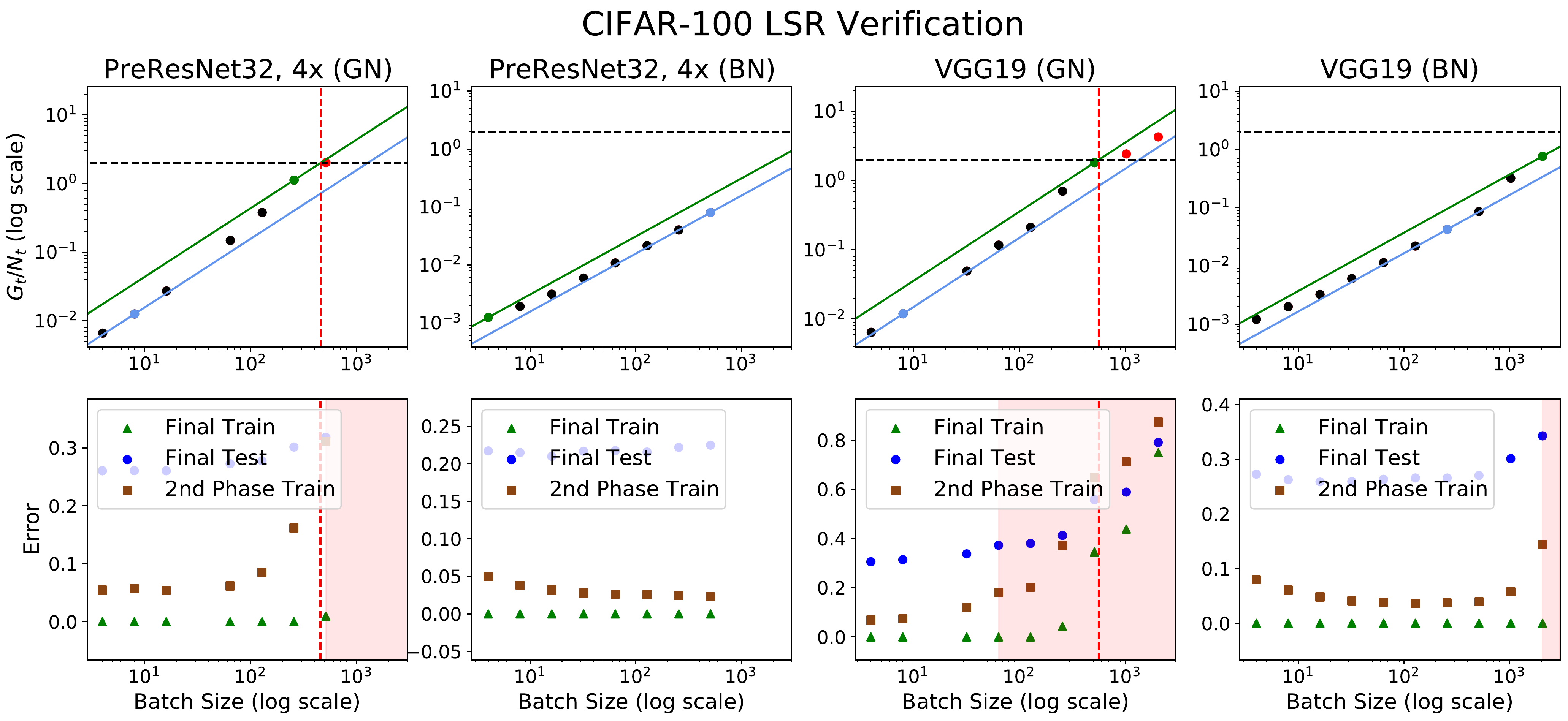}
    \caption{\small Further verification for our theory on predicting the failure of Linear Scaling Rule. We test if the condition applies to different architectures trained on CIFAR-100. All four settings use the same LR schedule, LR$=0.8$ initially and is decayed by $0.1$ at epoch $80$ and again at epoch $250$ with $300$ epochs total budget.  We measure $G_t$ and $N_t$ by averaging their values over the last 50 epochs of the second phase (i.e., from epoch 200 to 250).
    }
    \label{fig:app_lsr_c100}
\end{figure}

We further verify the necessary condition for LSR (\Cref{thm:lsr_kappa_bound}) using different architectures and datasets.
\Cref{fig:app_lsr_c10} tests the condition for ResNet-32 and wider PreResNets trained on CIFAR-10.
Although our theory requires strict scale-invariance, we find the condition to still be applicable to the standard ResNet architecture~\citep{he2016deep}, ResNet32, likely because most of the network parameters are scale-invariant.
\Cref{fig:app_lsr_c100} tests the condition for wider PreResNets and VGG-19 trained on CIFAR-100. 
We require the wider PreResNet to achieve reasonable test error, but we note that the larger model made it difficult to straightforwardly train with a larger batch size.  

In \Cref{fig:app_lsr_c10} and \Cref{fig:lsr_acc_gnr}, $G_t$ and $N_t$ are the empirical estimations of $\Gi$ and $\Ni$ taken after reaching equilibrium in the second to last phase (before the final LR decay), where the number of samples (batches) is equal to 
$\max(200, 50000/B)$, and $B$ is the batch size. 

     Per the approximated version of \Cref{thm:lsr_kappa_bound}, i.e., $B^*=\kappa B \lesssim C^2B{\Ni^B}/{\Gi^B}$, we use baseline runs with different batch sizes $B$ to report the maximal and minimal predicted critical batch size, defined as the x-coordinate of the intersection of the threshold ($\nicefrac{G_t}{N_t}=C^2$) with the green and blue lines, respectively. Both the green and blue line have slope $1$, and thus the x-coordinate of intersection, $B^*$, is the solution of the following equation, 
     \[\frac{B^*}{B} = \frac{G^{B^*}_t/N^{B^*}_t}{G^B_t/N^B_t}, \textrm{ where $G^{B^*}_t/N^{B^*}_t=C^2$.} \]
    For all settings, we choose a threshold of $C^2 = 2$, and consider LSR to fail if the final test error exceeds the lowest achieved test error by more than 20\% of its value, marked by the red region on the plot. Surprisingly, it turns out the condition in \Cref{thm:lsr_kappa_bound} is not only necessary, but also close to sufficient.

\begin{figure}[!htbp]
    \centering
    \includegraphics[width=0.5\linewidth]{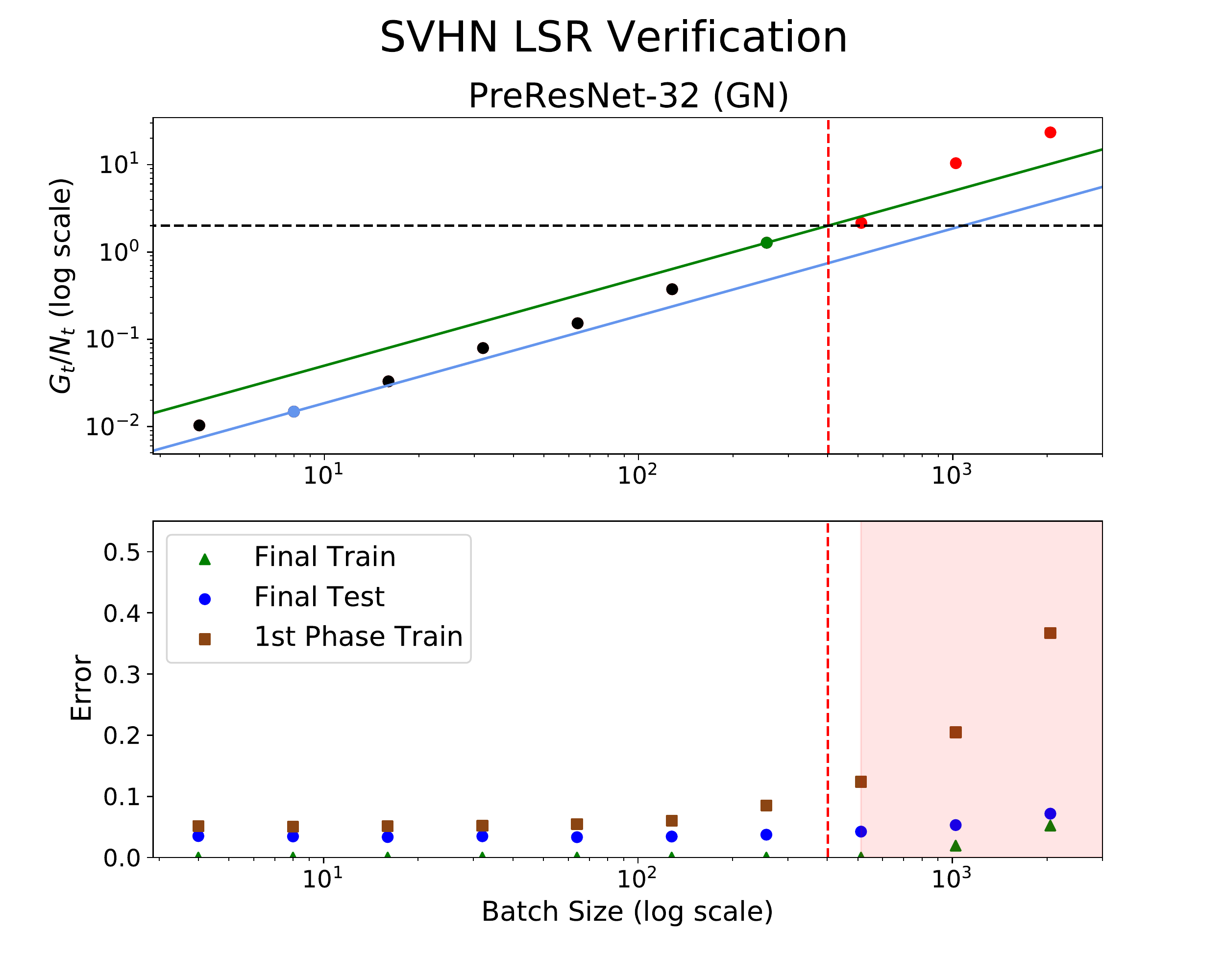}
    \caption{\small Further verification for our theory on predicting the failure of Linear Scaling Rule. We test if the condition applies to PreResNet-32 with GN trained on SVHN~\cite{netzer2011reading}. LR$=0.8$ initially and is decayed by $0.1$ at epoch $100$ with $120$ epochs total budget.  We measure $G_t$ and $N_t$ by averaging their values over the last 20 epochs of the second phase (i.e., from epoch 80 to 100).
    }
    \label{fig:app_lsr_svhn}
\end{figure}

\newpage

\subsection{Additional Experiments for NGD (Noisy Gradient Descent)}
\label{sec:app_ngd_exps}
\begin{figure}[!htbp]
    \centering
    \includegraphics[width=0.7\linewidth]{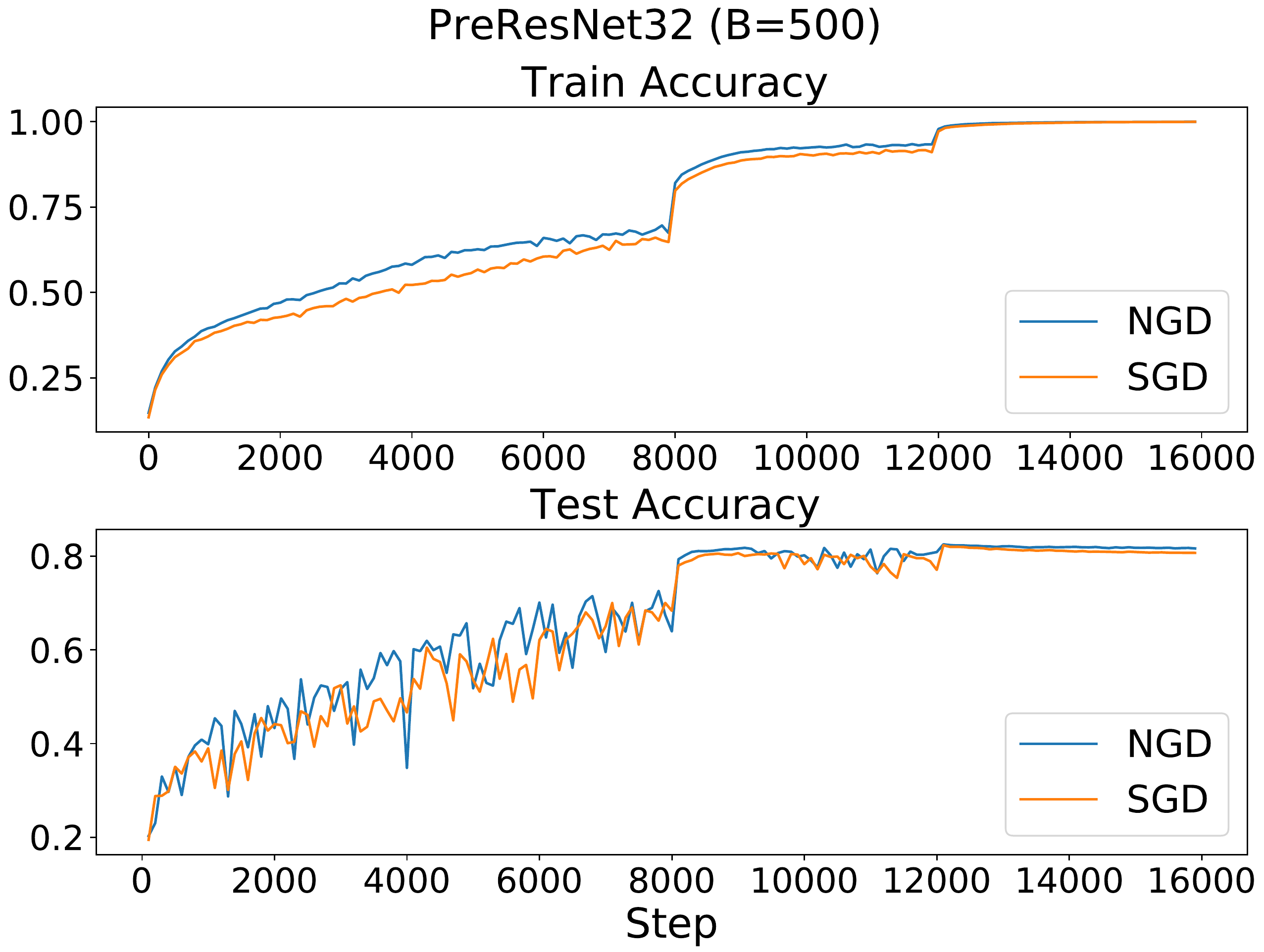}
    \caption{SGD and NGD with matching covariance have close train (top) and test (bottom) curves.  The batch size for SGD is $500$ and LR$=3.2$ for both settings and decayed by $0.1$ at step $8000$. We smooth the training curve by dividing it into intervals of 100 steps and recording the average. For efficient sampling of Gaussian noise, we use GroupNorm instead of BatchNorm and turn off data augmentation. SGD and NGD achieve a maximum test accuracy of 82.3\% and 82.5\%, respectively}    
    \label{fig:ngd_sgd_b500}
\end{figure}

\begin{figure}[!htbp]
    \centering
    \includegraphics[width=0.7\linewidth]{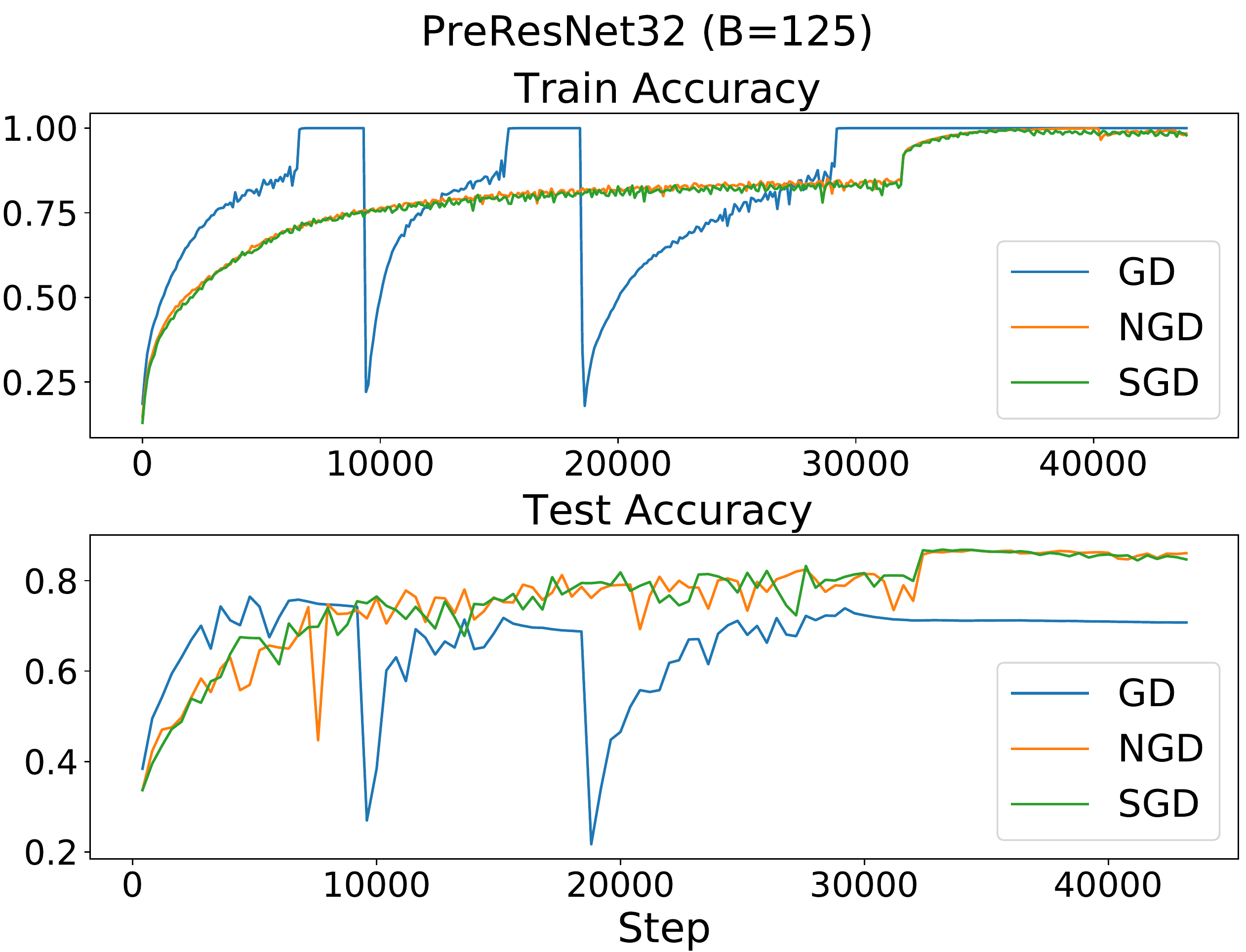}
    \caption{SGD and NGD with matching covariance have close train (top) and test (bottom) curves.  The batch size for SGD is $125$ and LR$=0.8$ for all three settings and decayed by $0.1$ at step $32000$. We smooth the training curve by dividing it into intervals of 100 steps and recording the average. For efficient sampling of Gaussian noise, we use GroupNorm instead of BatchNorm and turn off data augmentation. GD achieves a maximum test accuracy of 76.5\%, while SGD and NGD achieve 86.9\% and 86.8\%, respectively}
    \label{fig:ngd_sgd_b125}
\end{figure}

\begin{figure}[!htbp]
    \centering
    \includegraphics[width=0.7\linewidth]{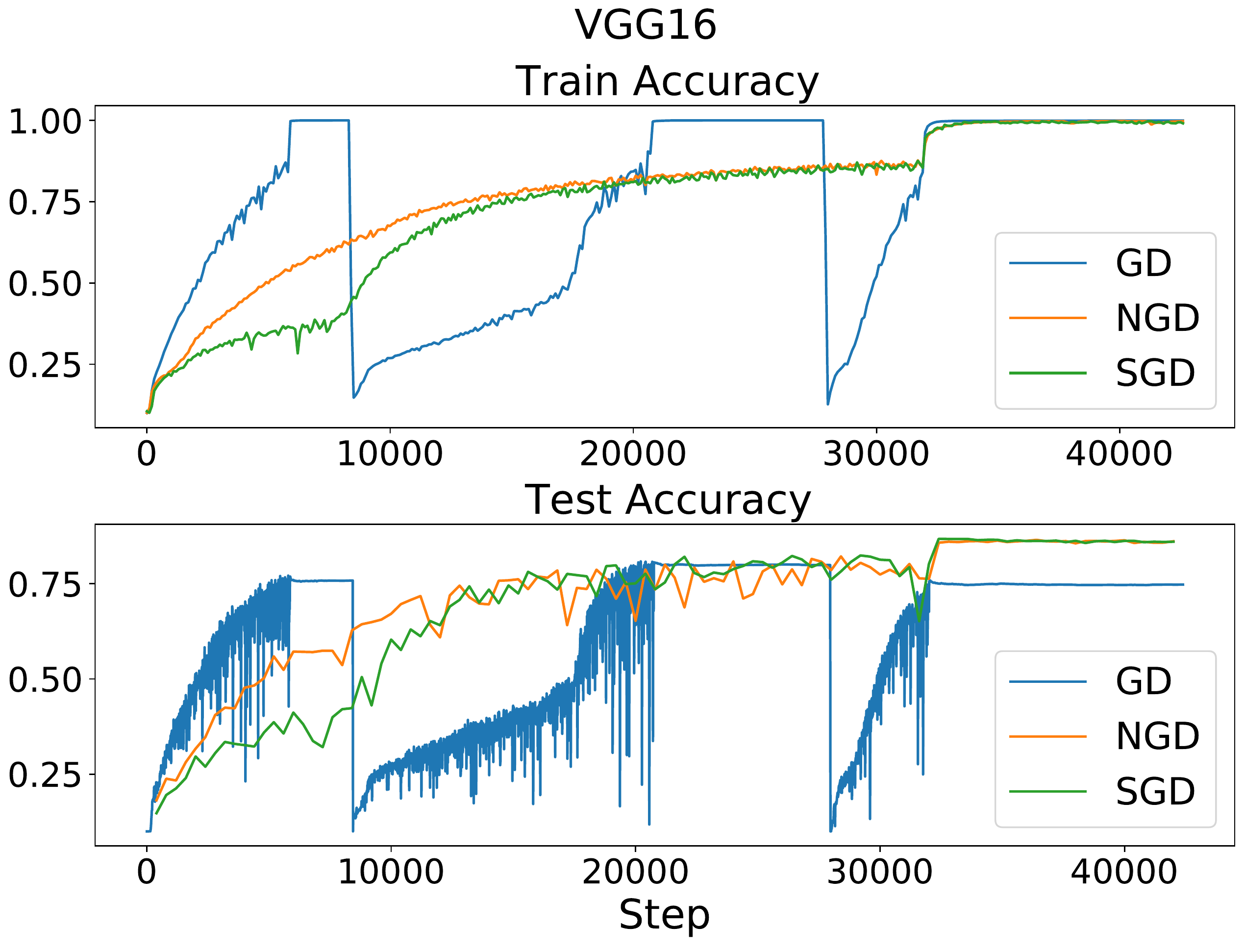}
    \caption{SGD and NGD with matching covariance have close train (top) and test (bottom) curves for VGG16.  The batch size for SGD is $125$ and LR$=0.8$ for all three settings and decayed by $0.1$ at step $32000$. We smooth the training curve by dividing it into intervals of 100 steps and recording the average. For efficient sampling of Gaussian noise, we use GroupNorm instead of BatchNorm and turn off data augmentation. GD achieves a maximum test accuracy of 80.9\%, while SGD and NGD achieve 86.8\% and 86.5\%, respectively}
    \label{fig:ngd_sgd_vgg}
\end{figure}

We provide further evidence that SGD~\eqref{eq:sgd_iter} and noisy gradient descent (NGD)~\eqref{eq:noisy_gd} have similar train and test curves in Figures \ref{fig:ngd_sgd_b500}, \ref{fig:ngd_sgd_b125}, and \ref{fig:ngd_sgd_vgg}.
To perform NGD, we replace the SGD noise by Gaussian noise with the same covariance as was done in \cite{wu2020noisy}. 
In \cite{wu2020noisy}, the authors trained a network using BatchNorm, which prevents the covariance of NGD from being exactly equal to that of SGD.
Hence, we use GroupNorm in our experiments, which improves NGD accuracy.
We note that each step of NGD requires computing the full-batch gradient over the entire dataset (in this case, done through gradient accumulation), which is much more costly than a single SGD step.
Each figure took roughly $7$ days on a single RTX 2080 GPU.

\end{document}